%% file: paper.tex
\ifdefined\XeTeXversion\else\pdfoutput=1\fi
\documentclass[10pt, logo, copyright]{nvidiatechreport}
\ifdefined\XeTeXversion\microtypesetup{tracking=false}\fi
\usepackage[authoryear,round]{natbib}
\usepackage{multirow,xspace,algpseudocode}
\input{math_commands.tex}

\definecolor{nvidiagreen}{RGB}{118,185,0}
\hypersetup{allcolors=nvidiagreen}
\makeatletter
\renewcommand{\abscontent}{%
    \par\vskip6pt
    {\noindent\bfseries\fontsize{12}{14}\selectfont Abstract\par}
    \vskip6pt
    \noindent
    \parbox{\dimexpr\linewidth}{\absfont \theabstract}
    \@ifundefined{@keywords}{}{
        \vskip1em \noindent \keywordsfont  Keywords: \@keywords}
}
\makeatother

\colorlet{DataGreen}{nvidiagreen!55!black}
\colorlet{StageBlue}{DataGreen}
\colorlet{StageLight}{nvidiagreen!5}
\colorlet{PanelLight}{nvidiagreen!5}
\colorlet{RuleGray}{nvidiagreen}
\colorlet{TableHead}{nvidiagreen!10}
\colorlet{VERAFill}{nvidiagreen!7}
\colorlet{VERAInk}{DataGreen}
\newcolumntype{Z}{>{\hsize=.8\hsize\linewidth=\hsize\centering\arraybackslash}X}
\newcolumntype{W}{>{\hsize=1.6\hsize\linewidth=\hsize\centering\arraybackslash}X}
\newcolumntype{Y}{>{\raggedright\arraybackslash}X}

\newcolumntype{C}{>{\centering\arraybackslash}p{0.185\linewidth}}

\newcommand{\autorubric}{\textsc{Auto-Rubric}\xspace}

\newcommand{\vera}{\textsc{VERA}\xspace}

\newcommand{\costCoworkTwentySevenB}{5{,}000}

\newcommand{\costMedTwentySevenB}{8{,}000}

\newcommand{\openenvs}{9{,}000+}

\newcommand{\cmark}{{\color{DeltaUp}\ding{51}}}%
\newcommand{\xmark}{{\color{DeltaDown}\ding{55}}}%

\definecolor{GoldMedal}{HTML}{C9A227}
\definecolor{SilverMedal}{HTML}{495057}
\newcommand{\goldmedal}{{\footnotesize\color{GoldMedal}\ding{72}}}
\newcommand{\silvermedal}{{\footnotesize\color{SilverMedal}\ding{72}}}

\definecolor{DeltaUp}{HTML}{129C4B}
\definecolor{DeltaDown}{HTML}{E03131}

\title{VERA: Scaling Verifiable Environments for Agentic co-Evolution}

\sauthor{2}{}{Junqi Liu}
\sauthor{3}{}{Yongyang Pan}
\sauthor{4}{}{Zhuosong Jiang}
\sauthor{5}{}{Dongbai Li}
\sauthor{5,6}{}{Bo Zhang}
\sauthor{5}{}{Xitong Ling}
\sauthor{7}{}{Sheng Wang}
\sauthor{1}{}{Hanrong Ye}
\sauthor{1}{}{Yufan He}
\sauthor{1}{}{Can Zhao}
\sauthor{1}{}{Pengfei Guo}
\sauthor{1}{}{Dong Yang}
\sauthor{1}{}{Andriy Myronenko}
\sauthor{2}{}{Yuyin Zhou}
\sauthor{5}{}{Tianyu Liu}
\sauthor{1}{}{Daguang Xu}
\sauthor{1}{*}{Yucheng Tang}

\saffil{1}{}{Nvidia}
\saffil{2}{}{University of California, Santa Cruz}
\saffil{3}{}{University of Illinois Urbana-Champaign}
\saffil{4}{}{National University of Singapore}
\saffil{5}{}{Tsinghua University}
\saffil{6}{}{Xian Jiaotong University}
\saffil{7}{}{University of Pennsylvania}

\spill{https://github.com/AutoMedBench/VERA}{Code}
\spill{https://automedbench.github.io/vera/}{Blog}

\begin{abstract}
Competent agents need precise and verifiable environments, such as sandboxes that are 
resumable at any stage and evolve from observable evidence. However, most long-horizon
work exposes how rare these are: for example, an agent in medical research must ground a
finding, classify it, and write a report over dozens of dependent steps, yet recent environments score only the outcome. To address the challenges in stable training, we present \vera{}, which builds such environments at scale and lets agents evolve on them. \vera{} builds these environments from initial trajectories: an agent writes rubrics, executable checks, a judge verifies each sandbox, and
only those that pass enter the training bank. On these environments, \vera{} alternates between two updates: train the model with rubric rewards, or edit the harness skills. We also create a verifier which gates model checkpoints and harness edits using explicit development-set acceptance criteria.
This attribution distinguishes \vera{}'s co-evolution from single-axis baselines: its updates target not only the cause but the outcome. With an open-source corpus of \openenvs{} long-horizon verifiable environments, a 9B model paired with its co-evolved agent beats the strongest baseline by 10.3 and 13.0 points in the two domains. At 27B, it surpasses the baseline on AutoCoWorkBench (71.6) and AutoMedBench (80.7), transfers to unseen workflows, and retains general capabilities. 

\end{abstract}

\begin{document}
\maketitle

\begin{figure}[!h]
    \centering
    \includegraphics[width=\linewidth]{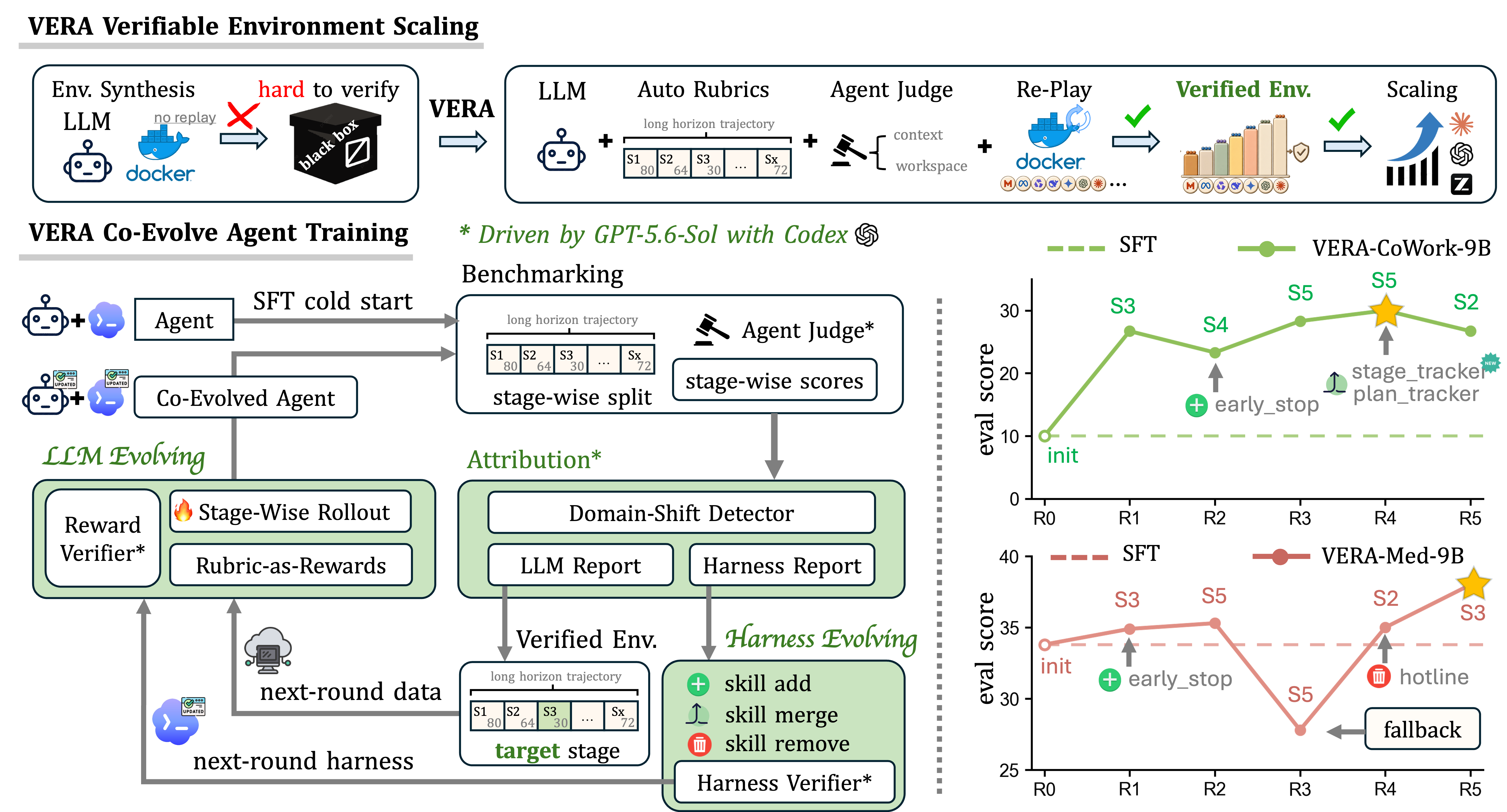}
    \caption{\textbf{VERA enables model--harness co-evolution through verifiable environments.}
    \textbf{Top:} Auto-generated rubrics, Agent Judge workspace checks,
    and container replay verify environments before scaling.
    \textbf{Bottom left:} Attribution uses stage-wise scores and traces
    to produce an LLM Report selecting training stages and a Harness
    Report proposing skill additions, merges, or removals.
    The Reward Verifier checks rubric rewards; the Harness Verifier
    tests skill edits. Updates alternate with the other component frozen,
    carrying accepted changes forward.
    \textbf{Right:} CoWork and MedResearch scores across rounds.
    Dashed lines denote SFT baselines; labels identify target stages
    and skill edits. Fallback restores a previously accepted checkpoint.}
    \label{fig:rlevo}
\end{figure}

\section{Introduction}
\label{sec:introduction}

One ultimate ambition of agentic artificial intelligence is a system that
improves from its own attempts and grows into an expert-level generalist.
AI has taken visible steps toward this ambition in software engineering
\citep{jimenez2024swebench,merrill2026terminalbench}, computer-use, office
automation \citep{yao2025taubench,zhou2024webarena}. Yet the ambition remains unrealized for the
long-horizon research. Consider medical research: a
competent research agent must ground findings in an image, classify them, and
compose a report under expert scrutiny, a workflow of dozens of
dependent steps in which one early mistake invalidates the entire workflow. Unlike coding and office workflows, where we can find trajectory resources online, very few AI systems today can learn to master research workflows because they lack practical environments. We argue that the bottleneck is not the learning algorithm but the
\emph{environment}. 

In modern agent training, an environment contains a reusable sandbox, a task packaged with its tools, workspace, and initial
state, together with a mechanism that scores success from observable evidence. Reinforcement Learning (RL) consumes such environments, and self-improvement is impossible without them. Current practice builds environments in two ways, and both break down on
long-horizon agentic work. First, manual construction
pipelines \citep{pan2024swegym,jain2025r2egym,yang2025swesmith} synthesize
executable tasks and define success by a test suite. Agents improve themselves through test signals, through RL with verifiable rewards \citep{shao2024deepseekmath,deepseek2025r1}, on-policy
distillation \citep{agarwal2023onpolicy}, or reflection training
\citep{bi2026reflectrl}; or through the harness, skills, tools, and execution
rules, via verbal reflection \citep{shinn2023reflexion}, skill libraries
\citep{wang2023voyager}, prompt and workflow search
\citep{khattab2024dspy,zhang2025aflow}, and harness-code evolution
\citep{hu2025adas,zhang2025dgm,lee2026metaharness}. Recent systems co-evolve
both \citep{cosplay2026,wu2026modularrsi}, but without verified evidence of a failure, each change is judged only by scores or pass rates. 
Second, evaluation benchmarks
\citep{jimenez2024swebench,merrill2026terminalbench,chan2024mlebench,%
schmidgall2024agentclinic,zuo2025medxpertqa} are also environments that are curated by hand and score only the outcome. This process does not measure rewards or localize failures in the intermediate stages. The result is a structural gap: environments yield no stage-level verified evidence, and self-evolution loops cannot gain from precision failures. In words, \emph{no precise verifiable environments}. 

In this paper, we present \vera{} (\emph{Verifiable Environments for Agentic co-Evolution}), which closes this gap by constructing open verifiable environments and letting the agent evolve on them. Our starting observation is that benchmark trajectories already contain everything a verifiable environment starts with: initial states, tool interactions, and their observable effects. A data pipeline, \autorubric, converts such trajectories into training sandboxes. A coding agent writes binary rubrics over action arguments, execution effects, artifacts, and outputs; stage-wise sandboxes restore the workspace at the start of each stage: \textit{plan}, \textit{setup}, \textit{verify}, \textit{execute}, \textit{submit}, so a long workflow decomposes into verifiable units, while end-to-end sandboxes use an LLM for tool calls \citep{zhang2026heddle}. Executable code checks and an agent judge inspect the workspace, and a failed executable check overrides any
judge score. On these environments, the \vera{} loop evolves the agent in alternating rounds: \emph{LLM Attribution} reads stage-wise rubric scores and execution traces to decide which step is failing; model rounds run full-parameter GRPO \citep{shao2024deepseekmath} with rubric rewards that a Reward Verifier checks against rollout evidence; harness rounds add, merge, or remove skills, accepted only when paired reruns with the frozen model confirm improvement. This design solves three problems: rewards come from what the agent actually did rather than from tests, long tasks are scored step by step, and
only one part of the agent changes at a time, so every gain has clear
evidence.

We evaluate \vera{} in two complementary domains: Medical Research and Automation. 
AutoMedBench~\citep{automedbench2026} and AutoCoWorkBench cover practical work and
specialized research workflows, respectively; both evaluate complete long-horizon tasks with process and outcome scores, allowing us to assess execution quality alongside task success. At 9B, co-evolution
outperforms the strongest single-axis baseline by 10.3 and 13.0 Overall points, respectively, with ablations confirming the importance of
updating both the model and its harness. At 27B, \vera{} reaches 71.6 on AutoCoWorkBench and 80.7 on AutoMedBench, surpassing Claude Opus 4.8
on the former and coming within 1.2 points on the latter. It also
transfers to unseen workflows on Automation-Bench~\citep{zapier2026automationbench} and AgentClinic~\citep{schmidgall2024agentclinic}.
Across 4B, 9B, and 27B backbones, domain performance increases across the tested backbones; only
at 27B are general capabilities also retained and mostly improved,
suggesting that stronger backbones better support self-evolution
without sacrificing broader capabilities.

To the best of our knowledge, we build the first large-scale verifiable
environments for the joint self-evolution of an LLM and its harness, and we
open-source the resulting \openenvs{} environment corpus.
To summarize, we make the following contributions:
\begin{itemize}
    \item \textbf{\vera{}: verifiable environment scaling for recursive
    self-improvement.} We turn benchmark trajectories into restartable,
    rubric-scored sandboxes for long-horizon agent tasks, with an
    admission loop that accepts only environments executable and scoreable
    from observable evidence.
    \item \textbf{Co-evolution of the LLM and its harness on
    \vera{} environments.} Rubrics provide verified evidence; LLM Attribution turns it
    into model and harness reports; the two components update in
    alternating rounds, and verifiers decide whether each update is kept.
    \item \textbf{An open resource of \openenvs{} verifiable long-horizon
    training environments and scaling to a 27B backbone.} At 9B the
    co-evolved agent exceeds the strongest single-axis baseline by 10.3
    (CoWork) and 13.0 (MedResearch) Overall points; at 27B it surpasses
    Claude Opus 4.8 on AutoCoWorkBench and comes within 1.2 points on
    AutoMedBench.
\end{itemize}

\begin{figure}[t]
    \centering
    \includegraphics[width=\linewidth]{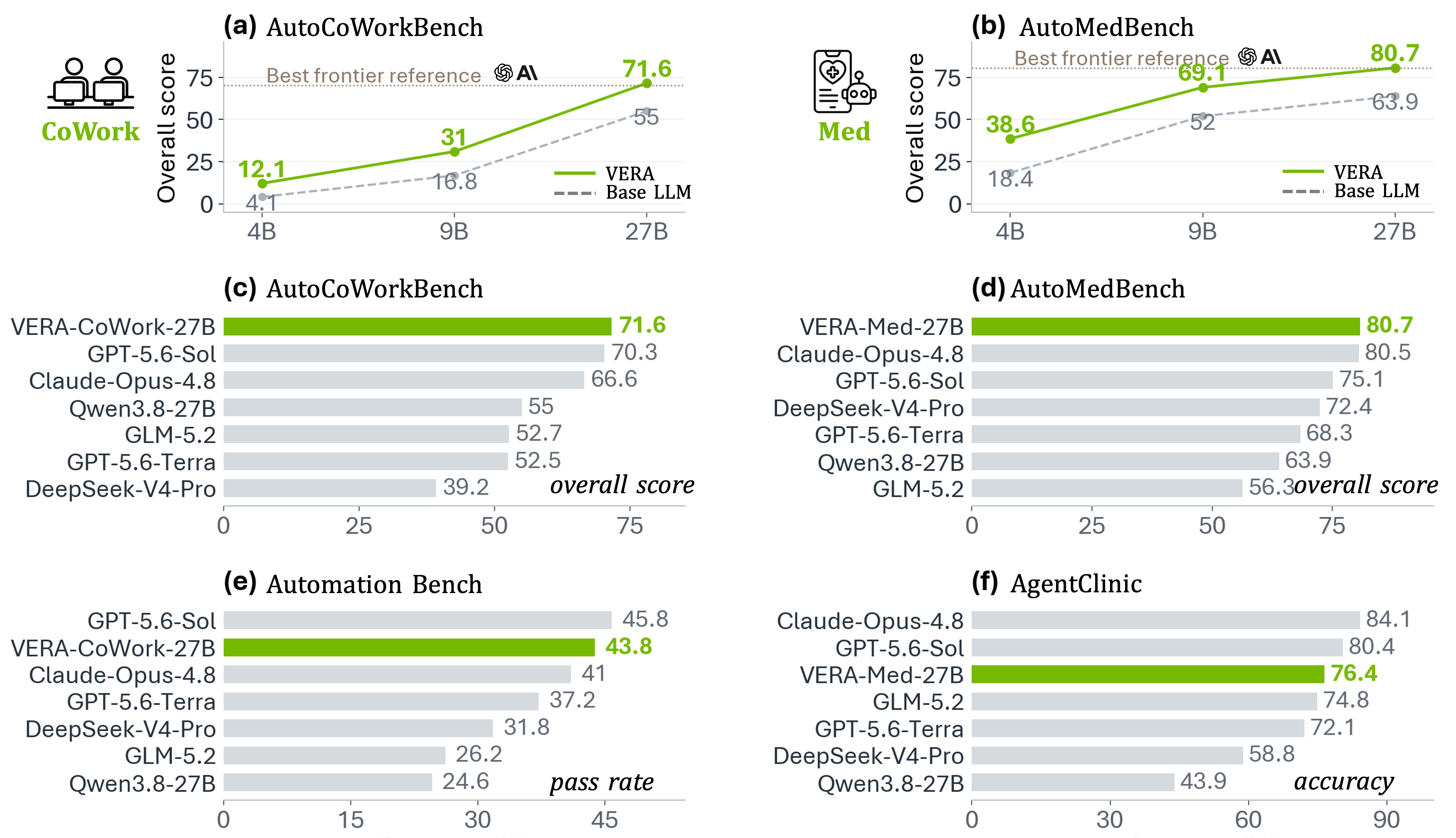}
    \caption{\textbf{VERA improves agents across model sizes and achieves competitive performance with a 27B backbone.}
    (a-b) Co-evolution improves on the base agents at 4B, 9B, and 27B;
    dotted lines mark the best frontier reference scores.
    (c-d) VERA-27B achieves the highest Overall scores among the
    compared agents on AutoCoWorkBench and AutoMedBench.
    (e-f) On Automation-Bench (public 600)
    \citep{zapier2026automationbench} and AgentClinic
    \citep{schmidgall2024agentclinic}, VERA-27B improves over its
    base agent by 19.2 and 32.5 percentage points, respectively,
    ranking second and third among the compared systems.
    These comparisons evaluate complete agents with their configured
    productivity harnesses, rather than model weights alone.
    Green denotes VERA; gray denotes comparison models.
    Harness and reasoning-effort settings appear in
    Appendix~\ref{app:harness_design} (Table~\ref{tab:ood_configs}).}
    \label{fig:scaling}
\end{figure}

\section{Method}
\label{sec:method}

\begin{figure}[t]
    \centering
    \includegraphics[width=\linewidth]{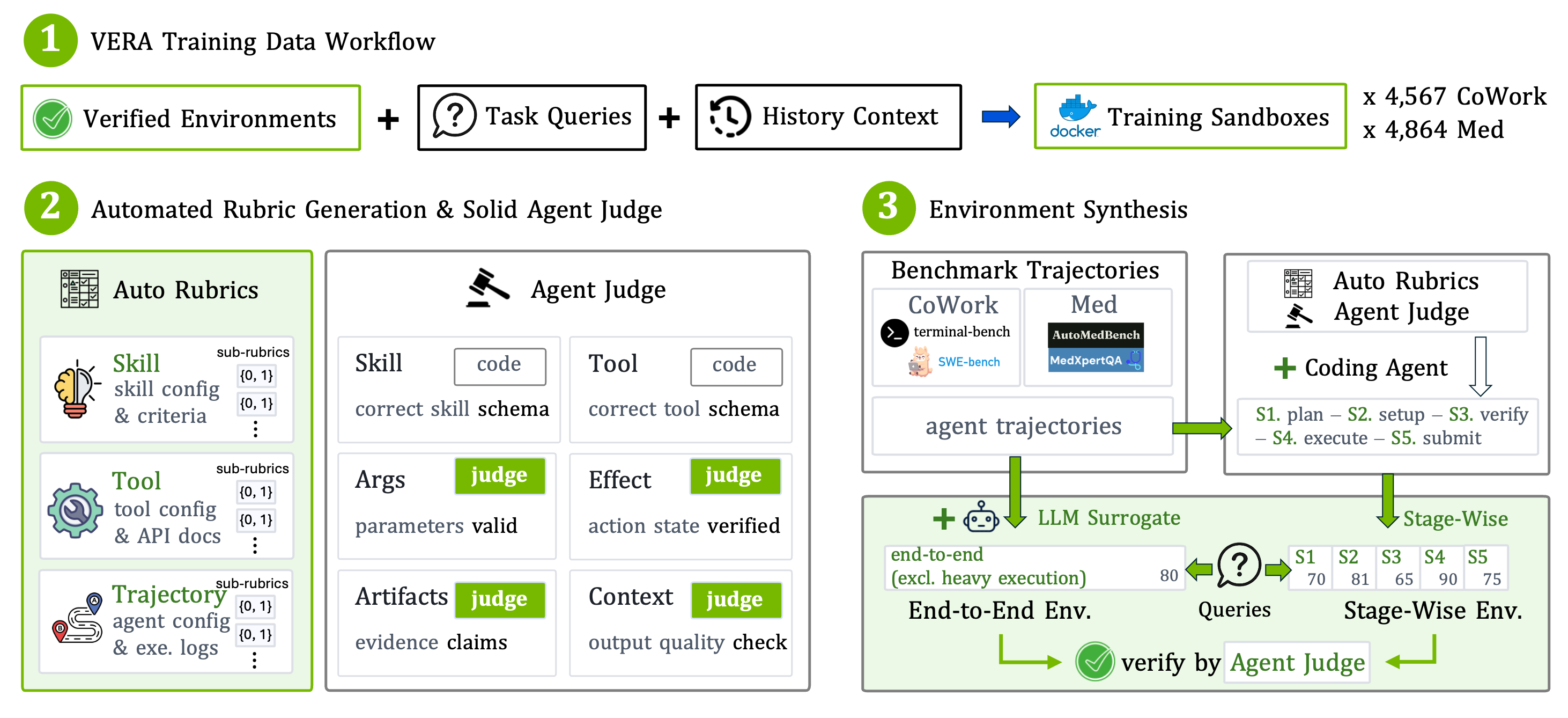}
    \caption{\textbf{\autorubric makes long workflows restartable and verifiable.}
    \textbf{(1) Training sandbox construction.}
    Each sandbox combines a verified environment (tools, workspace, and
    initial state), a task query, and the interaction history needed to
    resume the task. 
    \textbf{(2) Rubric generation and evidence verification.}
    A coding agent generates binary rubric items from skill configurations,
    tool API documentation, and execution logs. Each item defines a
    success criterion scored in $\{0,1\}$ from observable evidence.
    Executable checks validate skill and tool schemas.
    The Agent Judge
    inspects the trajectory and workspace to assess argument validity,
    actual execution effects, artifact evidence, and output quality in
    the task context. Failed executable checks override judge scores.
    \textbf{(3) Stage-wise and end-to-end environment synthesis.}
    Benchmark trajectories are segmented into five stages:
    plan, setup, verify, execute, and submit (S1--S5).
    Stage-wise environments restore each stage's starting state and attach
    the corresponding query, history, and rubrics, allowing that stage
    to be practiced and scored independently. End-to-end environments
    preserve the full workflow while an LLM surrogate simulates responses
    from time-consuming tool calls~\citep{ren2025gtm}. 
    Both types must pass executable checks and Agent Judge verification before entering the training bank.
    Numbers beside the environments illustrate rubric scores.}
    \label{fig:rlevo-data}
\end{figure}

We introduce \textbf{\vera{}} (\emph{Verifiable Environments for Agentic
co-Evolution}). Its goal is to let a long-horizon agent improve itself
from its own attempts, without human labels, through \emph{rule-guided
self-evolution}: verified rubrics provide the evidence, \emph{LLM
Attribution} turns the evidence into model-side and harness-side reports,
and verifiers decide whether each update is kept. Concretely, \vera{} consists of two pipelines
(Figure~\ref{fig:rlevo}): a \emph{data pipeline}, \autorubric, which turns
benchmark trajectories into verified, restartable training environments
(Figure~\ref{fig:rlevo-data}); and a \emph{training pipeline} that
co-evolves the LLM and its harness on these environments in alternating
rounds. The rest of this section
presents each pipeline in turn.

\subsection{\vera: Co-Evolution Method}
\label{sec:rlevo}

The \vera{} loop improves two parts of the agent in alternating rounds: the LLM
and the harness (its skills, tools, and execution rules).

Model training freezes the harness; harness testing freezes the model;
rubrics and verifiers stay fixed within a round. Any gain is therefore
attributable to a single axis, which is what lets rubric evidence decide
what to change next. To choose that axis, each round
starts by evaluating the current pair on development tasks. \emph{LLM
Attribution} reads the stage-wise scores and execution traces and
produces two reports: an \emph{LLM Report} that locates the target
stage, the stage with the most frequent model-attributed failures, and
a \emph{Harness Report} that lists missing or misleading procedures.
Rubrics provide the evidence; verifiers decide acceptance. Infrastructure
failures are excluded. Appendix~\ref{app:report_coevolution} illustrates
how the two reports direct model as well as skill updates in a CoWork example.

\paragraph{Update the model, freeze the harness.}
The model makes new attempts in the stage-wise and end-to-end sandboxes,
and full-parameter GRPO \citep{shao2024deepseekmath} updates its weights
with rubric scores as rewards, following the execution/training
separation of Agent Lightning \citep{luo2025agentlightning}. The
optimizer and budget match our GRPO baseline; only the learning signal
differs: rewards come from stage rubrics, and the curriculum follows the
target stage. For stage $k$, let $z_{k,j}\in\{0,1\}$ be
the score of rubric item $j$, $w_{k,j}>0$ its weight, and $h_k\in\{0,1\}$
indicate whether all mandatory checks pass:
\begin{equation}
r_k
=
h_k
\frac{\sum_j w_{k,j}z_{k,j}}
     {\sum_j w_{k,j}},
\qquad
R_{\mathrm{e2e}}
=
g\,[\lambda p + (1-\lambda)c],
\label{eq:stage_reward}
\end{equation}
where $p$ is a process score, $c\in\{0,1\}$ a verified completion
signal, $g\in\{0,1\}$ a trajectory-level hard gate, and $\lambda$ the
process weight. CoWork sets $p$ to the mean of the five stage rewards,
with $\lambda{=}0.2$ and $g{=}1$; MedResearch sets $p$ to a judged
process score, with $\lambda{=}0.5$ and $g{=}0$ upon verified reward
hacking or a hard-gate failure (Appendix~\ref{app:training_details}).
In words: a stage earns the weighted fraction of rubric items it passes,
and nothing if a mandatory check fails; the end-to-end reward gives
credit for partial progress but stays responsible for verified
completion. Before a reward enters training, the \emph{Reward Verifier}
checks the rollout evidence behind it. Training focuses on the target
stage from the LLM Report; neighboring stages and complete tasks stay in
the mix, so stage-level gains must survive the full workflow. A
candidate checkpoint that fails acceptance falls back to the previously
accepted checkpoint.
Unlike harness edits, model updates need not yield an immediate
improvement. A candidate checkpoint is accepted if its development-benchmark score
decreases by no more than 20\% relative to the previously accepted
checkpoint.

\paragraph{Update the harness, freeze the model.}
The Harness Report turns failed rubric items into candidate skill edits:
add a missing skill, merge overlapping ones, or remove a harmful one
(e.g., adding \texttt{early\_stop} or merging \texttt{stage\_tracker}
and \texttt{plan\_tracker} in Figure~\ref{fig:rlevo}).
The same LLM performs attribution and writes the proposed skill edits.
Harness updates are restricted to skills; tools and execution rules
remain fixed.
The \emph{Harness Verifier} tests each edit with paired runs on the
development set: the same frozen model on the same tasks, with and
without the edit, so only the edit can cause the difference. An edit is
accepted only if it improves its target stage, causes no end-to-end
regression, and passes safety checks; otherwise the current harness is
kept. At most three edits are accepted per round.
In addition, an edit must pass self-tests and improve the Overall
development score by at least 5 $\%$ points.
How often a skill is used is not evidence that it helps; only the paired
comparison counts. Across rounds, the two axes feed each other: accepted skills change the next round's training distribution, and the improved model exposes the
next harness weakness. Appendix~\ref{app:harness_acceptance_audit}
provides concrete skill proposals and verifier decisions.

\begin{table}[t]
    \centering
    \caption{\textbf{\vera{} jointly evolves the LLM and harness
    with rubric-based rewards and verification.}
    Weights and Harness indicate updated components.
    Reward and Verification indicate rubric-based training rewards
    and verified-rubric checks in our implementations.}
    \label{tab:signals}
    \begingroup
    \small
    \setlength{\tabcolsep}{3pt}
    \renewcommand{\arraystretch}{1.18}
    \begin{tabularx}{\linewidth}{@{}lYcccc@{}}
        \toprule
        \rowcolor{TableHead}
        \textbf{Evolution} &
        \textbf{Method} &
        \textbf{Weights} &
        \textbf{Harness} &
        \textbf{Reward} &
        \textbf{Verification} \\
        \midrule

        \multirow{2}{*}{Harness}
        & ADAS~\citep{hu2025adas}
        & \xmark & \cmark & \xmark & \xmark \\
        & \textcolor{VERAInk}{Ours (VERA Harness)}
        & \xmark & \cmark & \xmark & \cmark \\
        \midrule

        \multirow{6}{*}{LLM}
        & SFT~\citep{ouyang2022training}
        & \cmark & \xmark & \xmark & \xmark \\
        & GRPO~\citep{shao2024deepseekmath}
        & \cmark & \xmark & \cmark & \cmark \\
        & RaR~\citep{gunjal2025rubrics}
        & \cmark & \xmark & \cmark & \cmark \\
        & OPD~\citep{agarwal2023onpolicy}
        & \cmark & \xmark & \xmark & \xmark \\
        & ReflectRL~\citep{bi2026reflectrl}
        & \cmark & \xmark & \cmark & \cmark \\
        & \textcolor{VERAInk}{Ours (VERA LLM)}
        & \cmark & \xmark & \cmark & \cmark \\
        \midrule

        \multirow{2}{*}{Both}
        & COS-PLAY~\citep{cosplay2026}
        & \cmark & \cmark & \xmark & \xmark \\
        & \textcolor{VERAInk}{\textbf{Ours (VERA)}}
        & \cmark & \cmark & \cmark & \cmark \\
        \bottomrule
    \end{tabularx}
    \endgroup
\end{table}
\subsection{\autorubric}
\label{sec:auto_rubric}

\paragraph{Rubrics}
A coding agent reads the task statement, the tool documentation, the
execution logs, and the workspace, then writes a rubric: a list of binary
checks, each scored in $\{0,1\}$, over action arguments, execution
effects, artifacts, and output quality. Every check must be verifiable
from observable evidence, so a rubric is at once the completion condition
of a task and the evidence required to prove it.

\paragraph{Sandboxes}
As shown in Figure~\ref{fig:rlevo-data}, we construct CoWork sandboxes
from Terminal-Bench \citep{merrill2026terminalbench}, SWE-Bench
\citep{jimenez2024swebench}, MLE-Bench \citep{chan2024mlebench}, and medical sandboxes from AutoMedBench \citep{automedbench2026}
and MedXpertQA \citep{zuo2025medxpertqa} workflows.
Inspired by BaT's staged medical training \citep{bat2026benchmark},
we adopt five stages---\textit{plan}, \textit{setup}, \textit{verify},
\textit{execute}, and \textit{submit}---to explore transfer across domains.
A coding agent restores each stage's starting workspace, retains its
query and interaction history, and attaches domain- and stage-specific
rubrics grounded in task requirements and artifacts
(Appendix~\ref{app:stage_rubrics}).
Each stage can then be practiced and scored independently.
End-to-end sandboxes retain the full workflow while an LLM surrogate
simulates time-consuming tool calls \citep{ren2025gtm}.
Both types undergo executable checks and Agent Judge verification
before entering the training bank.

\paragraph{Verification}
Every candidate sandbox passes two kinds of checks. Executable code
checks cover tool and skill schemas, restored states, and deterministic
outputs. The Agent Judge, going beyond LLM-as-a-judge
\citep{zheng2023judging}, interacts with the workspace and checks
arguments, execution effects, artifacts, and output context. A failed
executable check or official verifier overrides any judge score.
Sandboxes that fail are revised or rejected; only environments that run
and score from evidence enter the training bank. This admission loop is
what makes the corpus, and later the rewards, resistant to reward
hacking. Appendix~\ref{app:verifier_audit} illustrates verifier decisions;
Appendix~\ref{app:post_training_data} shows the resulting SFT records,
RL starting states, and curation checks.

\begin{table}[t]
    \centering
    \caption{\textbf{\vera{} achieves the best Overall and Pass@1
    scores across both domains.}
    All methods use Qwen3.5-9B~\citep{qwen2026qwen359b}, with separate CoWork and MedResearch
    agents. O/A/T denote Overall/Agentic/Task, with
    Overall averaging Agentic and Task.
    SWE-Bench Verified \citep{jimenez2024swebench} and
    MedXpertQA Text \citep{zuo2025medxpertqa} report Pass@1.
    Bold and underlined values indicate the best and second-best
    scores per column, respectively.
    Configuration details appear in Appendix~\ref{app:full_table}.}
    \label{tab:main}
    \begingroup
    \footnotesize
    \setlength{\tabcolsep}{2.2pt}
    \renewcommand{\arraystretch}{1.18}
    \begin{tabularx}{\linewidth}{@{}>{\raggedright\arraybackslash}p{0.30\linewidth}*{3}{Z}W*{3}{Z}W@{}}
        \toprule
        \rowcolor{TableHead}
        \textbf{Method} &
        \multicolumn{3}{c}{\shortstack{\textbf{AutoCoWork}\\\textbf{Bench}}} &
        \shortstack{\textbf{SWE-Bench}\\\textbf{Verified}} &
        \multicolumn{3}{c}{\shortstack{\textbf{AutoMed}\\\textbf{Bench}}} &
        \shortstack{\textbf{MedXpertQA}\\\textbf{Text}} \\
        \rowcolor{TableHead}
        & \textbf{O} & \textbf{A} & \textbf{T} & \textbf{P@1}
        & \textbf{O} & \textbf{A} & \textbf{T} & \textbf{P@1} \\
        \midrule

        \multicolumn{9}{@{}l}{\underline{\textit{Base LLM + benchmark default harness}}} \\
        Qwen3.5-9B~\citep{qwen2026qwen359b}
        & 4.2 & 0.0 & 8.3 & 41.8
        & 19.9 & 11.3 & 28.4 & 35.4 \\
        Qwen3.5-9B~\citep{qwen2026qwen359b} + Prompt
        & 12.0 & 8.0 & 16.0 & 44.0
        & 22.1 & 35.3 & 8.9 & 40.8 \\
        \midrule

        \multicolumn{9}{@{}l}{\underline{\textit{Harness Evolving}}} \\
        ADAS \citep{hu2025adas}
        & \underline{20.7} & 22.3 & 19.1 & 52.4
        & 55.5 & 61.6 & \underline{49.5} & 42.0 \\
        \rowcolor{VERAFill}
        \textcolor{VERAInk}{\textbf{Ours} (VERA Harness)}
        & 18.2 & 21.3 & 15.1 & \underline{53.6}
        & \underline{56.1} & \underline{63.4} & 48.8 & 42.7 \\
        \midrule

        \multicolumn{9}{@{}l}{\underline{\textit{LLM Evolving}}} \\
        SFT \citep{ouyang2022training}
        & 6.7 & 1.7 & 11.7 & 42.0
        & 12.9 & 21.6 & 4.1 & 32.1 \\
        GRPO \citep{shao2024deepseekmath}
        & 5.8 & 5.0 & 6.7 & 14.8
        & 36.3 & 47.0 & 25.7 & 22.3 \\
        RaR \citep{gunjal2025rubrics}
        & 18.1 & 19.4 & 16.7 & 50.2
        & 31.9 & 39.2 & 24.5 & 42.8 \\
        OPD \citep{agarwal2023onpolicy}
        & 19.8 & 15.6 & \textbf{24.0} & 49.2
        & 41.2 & 51.8 & 30.6 & 44.5 \\
        ReflectRL \citep{bi2026reflectrl}
        & 17.3 & 20.1 & 14.5 & 48.4
        & 28.6 & 20.7 & 36.5 & 40.9 \\
        \rowcolor{VERAFill}
        \textcolor{VERAInk}{\textbf{Ours} (VERA LLM)}
        & 18.9 & \underline{23.6} & 14.2 & 51.2
        & 43.3 & 50.6 & 36.1 & \underline{45.3} \\
        \midrule

        \multicolumn{9}{@{}l}{\underline{\textit{Co-Evolving}}} \\
        COS-PLAY \citep{cosplay2026}
        & 19.9 & 18.2 & \underline{21.6} & 51.8
        & 26.4 & 33.8 & 18.9 & 30.4 \\
        \rowcolor{VERAFill}
        \textcolor{VERAInk}{\textbf{Ours} (\textbf{VERA Agent})}
        & \textbf{31.0} & \textbf{45.3} & 16.7 & \textbf{54.2}
        & \textbf{69.1} & \textbf{76.8} & \textbf{61.3} & \textbf{50.8} \\
        \bottomrule
    \end{tabularx}
    \endgroup
\end{table}

\section{Experimental Setup}
\label{sec:experiments}

We evaluate whether verified model--harness co-evolution improves
long-horizon agents in CoWork and MedResearch.
Our experiments compare learning methods, isolate each update axis,
and assess scaling, transfer, and general-capability retention.
\textbf{Domains and benchmarks.}
Each domain uses separate agents and training sandboxes, with
evaluation instances excluded from training.
CoWork evaluation uses AutoCoWorkBench and SWE-Bench Verified~\citep{jimenez2024swebench,openai2024sweverified};
MedResearch uses the current seven-track AutoMedBench-Lite release~\citep{automedbench2026,automedbench2026lite}
(denoted AutoMedBench throughout) and MedXpertQA Text~\citep{zuo2025medxpertqa}.
Automation-Bench (public 600, v1.0.6)~\citep{zapier2026automationbench} and AgentClinic~\citep{schmidgall2024agentclinic} assess transfer.
Appendix~\ref{app:benchmark_organization} describes the benchmark task organization and scoring units;
Appendix~\ref{app:evaluation_details} lists the evaluation protocols and harness settings.
\textbf{Comparisons and scaling.}
At 9B, we compare prompting, harness optimization, model training,
and joint evolution under shared training data and matched
fixed-harness RL budgets
(Appendix~\ref{app:baselines}).
Harness-only and LLM-only variants at 9B and 27B isolate each component.
We evaluate full co-evolution at 4B, 9B, and 27B, compare 27B agents
with frontier models, and measure capability retention on AIME 2026~\citep{maa2026aime},
ALFWorld~\citep{shridhar2021alfworld}, GPQA-Diamond~\citep{rein2023gpqa}, and IF-Bench~\citep{pyatkin2025ifbench}.
\textbf{Implementation details.}
We use Qwen3.5 at 4B and 9B~\citep{qwen2026qwen354b,qwen2026qwen359b} and Qwen3.8 at 27B~\citep{qwen2026qwen3827b}.
Each run starts with SFT, followed by alternating model and harness
updates. Training settings and execution configurations appear in
Appendices~\ref{app:training_details} and~\ref{app:evaluation_details}.

\begin{table}[t]
    \centering
    \caption{\textbf{VERA-27B comes within 1.2 Overall points of the
    best frontier baseline in each domain.}
    O/A/T denote Overall/Agentic/Task.
    Frontier baselines use benchmark-default runners, while
    VERA variants use the model and harness configurations
    specified by the row groups.
    This table complements Figure~\ref{fig:scaling}, which
    evaluates agents with productivity harnesses; rankings
    are specific to the evaluated configurations.
    Bold and underlined values indicate the best and second-best
    scores per column, respectively.
    Agent and harness configurations are detailed in
    Appendix~\ref{app:evaluation_details}.}
    \label{tab:product}
    \begingroup
    \small
    \setlength{\tabcolsep}{3pt}
    \renewcommand{\arraystretch}{1.18}
    \begin{tabularx}{\linewidth}{@{}YcccYccc@{}}
        \toprule
        \rowcolor{TableHead}
        \multicolumn{4}{c}{\textbf{AutoCoWorkBench}} &
        \multicolumn{4}{c}{\textbf{AutoMedBench}} \\
        \rowcolor{TableHead}
        \textbf{Agent} &
        \textbf{O} & \textbf{A} & \textbf{T} &
        \textbf{Agent} &
        \textbf{O} & \textbf{A} & \textbf{T} \\
        \midrule

        \multicolumn{8}{@{}l}{\underline{\textit{Base LLM + benchmark default harness}}} \\
        Qwen3.8-27B~\citep{qwen2026qwen3827b}
        & 63.8 & 60.8 & 66.7 &
        Qwen3.8-27B~\citep{qwen2026qwen3827b}
        & 65.1 & 81.6 & 48.6 \\
        GPT-5.6-Sol~\citep{openai2026gpt56}
        & \textbf{72.6} & \textbf{76.8} & \underline{68.3} &
        GPT-5.6-Sol~\citep{openai2026gpt56}
        & 75.1 & 80.1 & \underline{70.0} \\
        Claude-Opus-4.8~\citep{anthropic2026opus48}
        & 66.6 & 70.7 & 62.5 &
        Claude-Opus-4.8~\citep{anthropic2026opus48}
        & \textbf{81.9} & 88.1 & \textbf{75.8} \\
        \midrule

        \multicolumn{8}{@{}l}{\underline{\textit{Base LLM + VERA Harness}}} \\
        \textcolor{VERAInk}{Qwen3.8-27B~\citep{qwen2026qwen3827b} + VERA-Code}
        & 66.4 & 70.1 & 62.7 &
        \textcolor{VERAInk}{Qwen3.8-27B~\citep{qwen2026qwen3827b} + VERA-Med}
        & 54.2 & 70.1 & 38.3 \\
        \midrule

        \multicolumn{8}{@{}l}{\underline{\textit{VERA LLM + benchmark default harness}}} \\
        \textcolor{VERAInk}{VERA-CoWork-27B}
        & 69.6 & 73.2 & 65.9 &
        \textcolor{VERAInk}{VERA-Med-27B}
        & 75.4 & \underline{91.7} & 59.2 \\
        \midrule

        \multicolumn{8}{@{}l}{\underline{\textit{VERA LLM + VERA Harness}}} \\
        \textcolor{VERAInk}{\textbf{VERA-CoWork-27B Agent}}
        & \underline{71.6} & \underline{73.7} & \textbf{69.4} &
        \textcolor{VERAInk}{\textbf{VERA-Med-27B Agent}}
        & \underline{80.7} & \textbf{94.1} & 67.4 \\
        \bottomrule
    \end{tabularx}
    \endgroup
\end{table}

\section{Results}
\label{sec:results}

\subsection{\vera{} Agentic Co-Evolution Outperforms Single-Axis Optimization}
\label{sec:main_results}

\vera{} improves agent performance across both domains by jointly
evolving the model and its harness (Table~\ref{tab:main}).
It outperforms harness evolving with ADAS~\citep{hu2025adas},
LLM evolving training including GRPO~\citep{shao2024deepseekmath}
and RaR~\citep{gunjal2025rubrics}, and harness and llm joint evolution with
COS-PLAY~\citep{cosplay2026}.
On AutoCoWorkBench, VERA-Code Agent reaches 31.0 Overall,
exceeding the strongest single-axis baseline, ADAS (20.7),
by 10.3 points.
On AutoMedBench~\citep{automedbench2026}, VERA-Med Agent reaches
69.1, exceeding its harness-only variant (56.1) by 13.0 points.
The full method also leads on SWE-Bench Verified
\citep{jimenez2024swebench} and MedXpertQA Text
\citep{zuo2025medxpertqa}, with 54.2 and 50.8 Pass@1, respectively.
These suites use official scoring protocols that differ from
the training rubrics.
Together, these results support the effectiveness of verified
co-evolution across both stage-scored workflows and final-task
evaluations.

\paragraph{Robustness}
\label{sec:robust_domains}

The same loop drives both lines and improves both: one method, two
tool ecosystems, no cross-domain interference, see \tableautorefname{}~\ref{tab:main}. The robustness comes
from the environments themselves, the same verifiable, restartable
sandboxes feed the loop in both domains. The lines improve along
different axes, harness and model contribute roughly equally on CoWork,
while the harness arm carries more of the Med gain, so a full
assessment needs the axis-level ablation and the scale analysis below.

\paragraph{Ablation Study.}
Freezing one axis costs about half the gain, see \tableautorefname{}~\ref{tab:main}. On AutoMedBench, LLM-only
reaches 43.3 and harness-only 56.1, against 69.1 for the full loop; on
AutoCoWorkBench, 18.9 and 18.2 against 31.0.

\subsection{\vera{} Scales to 27B and Reaches Frontier-Level Performance}
\label{sec:scaling_results}

Figure~\ref{fig:scaling} compares complete agents with productivity
harnesses, where \vera{} achieves the highest Overall scores on
AutoCoWorkBench and AutoMedBench.
Table~\ref{tab:product} evaluates base and frontier models with
benchmark-default runners instead.
Under these settings, \vera{} scores 71.6 on AutoCoWorkBench,
1.0 point below GPT-5.6-Sol, and 80.7 on
AutoMedBench~\citep{automedbench2026}, 1.2 points below Claude-Opus-4.8.
Claude scores 80.5 with its productivity harness but 81.9 with
the benchmark-default runner, explaining the different rankings.
This highlights the importance of harness configuration in
long-horizon tasks.
Full co-evolution outperforms both single-axis variants in Overall
across both domains and achieves the highest AutoMedBench Agentic
score (94.1).

We assess out-of-distribution generalization on Automation-Bench
and AgentClinic, neither of which is used in sandbox construction.
\vera{} achieves 43.8\% pass rate on
Automation-Bench~\citep{zapier2026automationbench}, ranking second,
and 76.4\% accuracy on AgentClinic~\citep{schmidgall2024agentclinic},
ranking third among the compared agents
(Figure~\ref{fig:scaling}).

\subsection{Stronger Backbones Improve Agent Performance}
\label{sec:llm_scaling}

Across AutoCoWorkBench and AutoMedBench~\citep{automedbench2026},
\vera{} improves over its backbone at every tested scale,
with Overall scores increasing from 4B to 9B to 27B
(Figure~\ref{fig:scaling}). This comparison reflects both size
and generation: Qwen3.5 supplies the 4B and 9B backbones,
and Qwen3.8 supplies 27B.

General-capability retention varies by scale
(Table~\ref{tab:retention}). The 4B and 9B agents decline on
several benchmarks, with AIME losses of 56.4-59.7 and
15.8-55.8 points, respectively. At 27B, the CoWork agent
improves on all four benchmarks: AIME 2026,
ALFWorld~\citep{shridhar2021alfworld},
GPQA-Diamond~\citep{rein2023gpqa}, and
IF-Bench~\citep{pyatkin2025ifbench}.
The Med agent improves on AIME and GPQA-Diamond, losing
4.0 points on ALFWorld and 0.7 on IF-Bench.
Thus, 27B provides the strongest domain performance and
capability retention among the tested backbones.

\begin{table}[t]
    \centering
    \caption{\textbf{General-capability retention.}
    Parentheses show score changes from the corresponding backbone
    (green: gains; red: drops).
    }
    \label{tab:retention}
    \begingroup
    \small
    \setlength{\tabcolsep}{3pt}
    \renewcommand{\arraystretch}{1.18}
    \newcommand{\retgain}[1]{\,{\scriptsize\textcolor{DeltaUp}{(+#1)}}}
    \newcommand{\retloss}[1]{\,{\scriptsize\textcolor{DeltaDown}{($-$#1)}}}
    \begin{tabularx}{\linewidth}{@{}Ycccc@{}}
        \toprule
        \rowcolor{TableHead}
        \textbf{Model} &
        \textbf{AIME 2026} &
        \textbf{ALFWorld} &
        \textbf{GPQA-Diamond} &
        \textbf{IF-Bench} \\
        \midrule
        Qwen3.5-4B~\citep{qwen2026qwen354b}
        & 89.7 & 32.1 & 76.2 & 59.2 \\
        \textcolor{VERAInk}{VERA-CoWork-4B}
        & 33.3\retloss{56.4} & 38.1\retgain{6.0}
        & 64.3\retloss{11.9} & 65.5\retgain{6.3} \\
        \textcolor{VERAInk}{VERA-Med-4B}
        & 30.0\retloss{59.7} & 18.2\retloss{13.9}
        & 61.1\retloss{15.1} & 34.5\retloss{24.7} \\
        \midrule
        Qwen3.5-9B~\citep{qwen2026qwen359b}
        & 92.5 & 74.6 & 81.7 & 66.7 \\
        \textcolor{VERAInk}{VERA-CoWork-9B}
        & 36.7\retloss{55.8} & 53.2\retloss{21.4}
        & 57.1\retloss{24.6} & 59.8\retloss{6.9} \\
        \textcolor{VERAInk}{VERA-Med-9B}
        & 76.7\retloss{15.8} & 68.4\retloss{6.2}
        & 81.2\retloss{0.5} & 52.0\retloss{14.7} \\
        \midrule
        Qwen3.8-27B~\citep{qwen2026qwen3827b}
        & 86.7 & 94.0 & 82.2 & 79.5 \\
        \textcolor{VERAInk}{VERA-CoWork-27B}
        & 96.7\retgain{10.0} & 97.8\retgain{3.8}
        & 89.9\retgain{7.7} & 81.3\retgain{1.8} \\
        \textcolor{VERAInk}{VERA-Med-27B}
        & 90.0\retgain{3.3} & 90.0\retloss{4.0}
        & 87.8\retgain{5.6} & 78.8\retloss{0.7} \\
        \bottomrule
    \end{tabularx}
    \endgroup
\end{table}

\subsection{\vera{} RSI Cost Analysis}
\label{app:rsi_cost}

\begin{figure}[ht]
    \centering
    \includegraphics[width=0.6\linewidth]{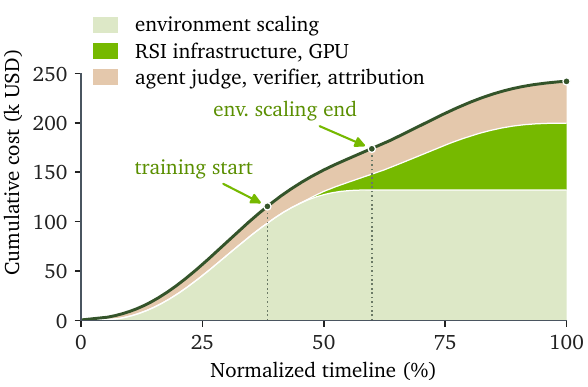}
    \caption{\textbf{Environment preparation dominates the estimated
    RSI cost in Medical Research.}
    Stacked areas show environment scaling and curation,
    training infrastructure, and verification and attribution costs.
    Dashed lines mark the start of model-harness co-evolution
    at 75\% of environment preparation and the completion of environment preparation.
    The normalized timeline and smooth curves illustrate the schedule rather than measured spending over time.}
    \label{fig:cost_scaling}
\end{figure}

We analyze the estimated cost of recursive self-improvement (RSI) in Medical Research pipeline.
Figure~\ref{fig:cost_scaling} illustrates the cumulative allocation, totaling approximately \$240k.

\paragraph{Cost breakdown.}
Environment scaling and curation account for 54.7\% of the total, covering sandbox construction, data curation, execution validation, and rubric verification.
Training infrastructure accounts for 27.9\%, while verification, judging, attribution, and supporting orchestration account for 17.3\%. Environment preparation is therefore the largest cost component in this pipeline.

\paragraph{Timeline and Scope.}
As illustrated in Figure~\ref{fig:cost_scaling},
model-harness co-evolution begins after approximately 75\% of environment preparation, allowing training and curation to overlap. 
The curves use a normalized, illustrative schedule rather than measured spending records.
Costs represent a replacement-cost estimate for this Medical Research pipeline, rather than audited expenditure or a general cost scaling law.

\section{Related Work}
\label{sec:related_work}

\subsection{Recursive Self-Improvement: Weights, Harness, or Both}
Self-improvement requires reliable feedback \citep{liu2026agentic}. On the \emph{weights}
side, GRPO uses group-relative rewards \citep{shao2024deepseekmath},
on-policy distillation learns from teacher feedback on student rollouts
\citep{agarwal2023onpolicy}, and ReflectRL learns from negative
trajectories \citep{bi2026reflectrl}. Agent Lightning separates execution
from RL training \citep{luo2025agentlightning}, while
Benchmark-as-Teacher \citep{bat2026benchmark} uses medical benchmark
rubrics \citep{automedbench2026} for stage-wise GRPO training.
On the \emph{harness} side, methods improve reflection
\citep{shinn2023reflexion}, skill libraries \citep{wang2023voyager},
prompts and pipelines \citep{khattab2024dspy,agrawal2025gepa},
or agent workflows \citep{zhang2025aflow,hu2025adas}.
Meta-Harness searches harness code using prior candidates and traces
\citep{lee2026metaharness}; the Darwin G\"odel Machine evolves
self-modifying agents \citep{zhang2025dgm}; concurrent work studies
harness-level RSI \citep{wu2026modularrsi}.
COS-PLAY jointly evolves a decision agent and skill bank
\citep{cosplay2026}.
Our focus is evidence-guided coordination of these two axes:
\vera{} uses stage-wise rubric evidence to attribute failures,
alternates model and harness updates, and checks acceptance on a
development set. We release \openenvs{} training sandboxes to
support further RSI research.

\subsection{Verifiable Rewards and Judges}
Reinforcement learning with verifiable rewards replaces a learned reward
model with a programmatic checker, and has driven recent gains in
reasoning \citep{lambert2024tulu3,deepseek2025r1}; process supervision adds
verifiable intermediate signals along the solution path
\citep{lightman2023verify,wang2024mathshephered}. Outside domains with exact
answers, rubrics and checklists break open-ended quality into itemized
criteria that can serve directly as rewards
\citep{gunjal2025rubrics,viswanathan2025checklists}. Judging has evolved
along a similar path, from LLM-as-a-judge \citep{zheng2023judging} to
agent-as-a-judge, where an agent inspects the workspace and intermediate
steps of another agent instead of reading only its final answer
\citep{zhuge2024agentjudge}. Any reward produced by a model, however,
can be over-optimized by the policy it trains
\citep{gao2023overoptimization}. Existing rubric rewards score single
responses, and agent judges have so far been used for evaluation rather
than for training. \vera{} writes rubrics per stage of a long-horizon
trajectory, grounds each check in an agent judge that inspects the
workspace, and lets a failed executable check override any model
judgment, so the same rubrics can serve as both an admission gate and a
training reward.

\subsection{Environments for Training Agents}
Agent benchmarks now ship realistic task suites with official checkers:
SWE-Bench Verified \citep{jimenez2024swebench}, Terminal-Bench
\citep{merrill2026terminalbench}, MLE-Bench \citep{chan2024mlebench},
$\tau$-Bench \citep{yao2025taubench}, and domain suites for web
\citep{zhou2024webarena} and medical
\citep{schmidgall2024agentclinic,zuo2025medxpertqa} agents. These are
static evaluation sets: they score the final outcome of a full episode and
cannot be restarted at an intermediate stage. A parallel line scales
executable environments for training coding agents, collecting or
synthesizing repository tasks at scale
\citep{pan2024swegym,jain2025r2egym,yang2025swesmith}; success there is
defined by the repository's test suite, which ties the approach to domains
with tests and rewards only the end of an episode. \vera{} instead converts
benchmark trajectories from both coding and medical research into sandboxes
that restart at any of five stages, attaches a verified rubric to each
stage, and admits a sandbox only if it runs and scores from observable
evidence \citep{chen2026ai}.

\section{Limitations}
\label{sec:limitations}
Our study leaves five aspects for further investigation.
\textbf{(1) Environment scaling:} We do not vary environment counts or
compare construction methods.
\textbf{(2) Harness ablations:} Runtime settings, including reasoning
effort, are fixed and not ablated.
\textbf{(3) Judge dependence:} Judging, verification, and attribution rely
on GPT-5.6-Sol; executable checks constrain its judgments, but
sensitivity to judge choice remains untested.
\textbf{(4) Baseline scale:} Full baseline comparisons are limited to 9B;
a 27B co-evolution run costs approximately
\costCoworkTwentySevenB{}-\costMedTwentySevenB{} GPU-hours.
\textbf{(5) Comparison scope:}
We compare learning algorithms rather than alternative operators
within VERA. Rubrics-as-Rewards remains fixed in our loop;
we do not separately ablate alternative reward formulations or
update operators.

\section{Conclusion}
\label{sec:conclusion}

We introduced \vera{}, a framework for recursive self-improvement
through verifiable environments and model-harness co-evolution.
\autorubric converts task trajectories into restartable sandboxes
with evidence-grounded feedback. This feedback identifies failures,
guides alternating model and skill updates, and supports verification
of proposed changes.

Across CoWork and MedResearch, co-evolution outperforms updating either component alone, with further gains at larger model scales.
By releasing \openenvs{} verifiable environments, we provide a foundation for studying how agents can learn from their own attempts.
Future work can extend this approach to broader workflows and examine how environment scale and verification quality affect sustained self-improvement.

\bibliography{paper}
\bibliographystyle{plainnat}

\appendix

\clearpage

\section*{Appendix Contents}
\begingroup
\setlength{\parindent}{0pt}
\setlength{\parskip}{3pt}
\newcommand{\appcontentsentry}[3]{%
  \noindent\hspace*{#1}\hyperref[#2]{\ref*{#2}\quad #3}%
  \nobreak\dotfill\nobreak\hyperref[#2]{\pageref*{#2}}\par}

\appcontentsentry{0pt}{app:benchmark_organization}{Benchmark Organization}
\appcontentsentry{1.5em}{app:autocoworkbench}{AutoCoWorkBench}
\appcontentsentry{1.5em}{app:automedbench}{AutoMedBench}

\appcontentsentry{0pt}{app:evaluation_details}{Evaluation Protocols and Agent Harness}
\appcontentsentry{1.5em}{app:harness_design}{Evaluation Harness Design}

\appcontentsentry{0pt}{app:stage_rubrics}{Verifiable Environments and Rubrics}
\appcontentsentry{1.5em}{app:stage_structure}{Stage Structure}
\appcontentsentry{1.5em}{app:cowork_rubrics}{CoWork Rubrics}
\appcontentsentry{1.5em}{app:med_rubrics}{MedResearch Rubrics}
\appcontentsentry{1.5em}{app:rubric_validity}{Scoring Scope and Evidence Validity}

\appcontentsentry{0pt}{app:post_training_data}{Post-Training Data Example}
\appcontentsentry{1.5em}{app:data_example_sft}{SFT}
\appcontentsentry{1.5em}{app:data_example_rl}{RL}
\appcontentsentry{1.5em}{app:data_example_curation}{Data Curation}

\appcontentsentry{0pt}{app:training_details}{VERA Training Details}
\appcontentsentry{1.5em}{app:hparams}{RL Hyperparameters}
\appcontentsentry{1.5em}{app:reward_details}{Reward}
\appcontentsentry{1.5em}{app:figure_walkthrough}{Co-Evolution Components}

\appcontentsentry{0pt}{app:baselines}{Baseline Configurations}
\appcontentsentry{1.5em}{app:baseline_configs}{Compared Configurations and Shared Protocol}
\appcontentsentry{1.5em}{app:sft_initializer}{SFT Initialization}
\appcontentsentry{1.5em}{app:harness_search}{ADAS Adaptation}
\appcontentsentry{1.5em}{app:cosplay_reproduction}{COS-PLAY Adaptation}
\appcontentsentry{1.5em}{app:full_table}{Full CoWork and MedResearch Results}

\appcontentsentry{0pt}{app:report_coevolution}{Report-Driven Co-Evolution Example}

\appcontentsentry{0pt}{app:verifier_audit}{Verifier Case Studies}

\appcontentsentry{0pt}{app:harness_acceptance_audit}{Harness Case Studies}

\endgroup

\clearpage

\section{Benchmark Organization}
\label{app:benchmark_organization}

AutoCoWorkBench and AutoMedBench evaluate complete workflows and expose
both outcome and process scores. Their evaluation units differ: a CoWork
episode solves one adapted task, whereas a medical workflow produces
predictions or artifacts for a track's collection of cases. The benchmark
inventories below describe the released task organization; the experiment
configurations in Appendix~\ref{app:evaluation_details} specify the agents
and harnesses used in this report.

\subsection{AutoCoWorkBench}
\label{app:autocoworkbench}

\paragraph{Evaluation tasks and supporting pools.}
The AutoCoWorkBench Lite core contains 40 end-to-end tasks: 15 repository
repairs adapted from SWE-bench Verified, 10 Terminal-Bench tasks, and 15
\mbox{tau2} interactions spanning airline, retail, and telecom
workflows.
Table~\ref{tab:benchmark_boundaries} separates this evaluation core from
the repository's supporting pools. The standalone SWE-Bench Verified
column in our results uses the official SWE-bench protocol; the adapted
15-task component does not replace that evaluation.

\begin{table}[!ht]
\centering
\caption{\textbf{AutoCoWorkBench separates evaluation tasks from training and gating pools.}
Counts refer to the dataset repository's task manifests.}
\label{tab:benchmark_boundaries}
\begingroup
\small
\setlength{\tabcolsep}{4pt}
\renewcommand{\arraystretch}{1.12}
\begin{tabularx}{\linewidth}{@{}p{.24\linewidth}p{.12\linewidth}Y@{}}
\toprule
\rowcolor{TableHead}
\textbf{Collection} & \textbf{Size} & \textbf{Role and boundary} \\
\midrule
Lite evaluation core & 40 tasks & Complete episodes; the official Lite evaluation denominator. \\
Training stage pool & 56 tasks & Stage-specific practice; excluded from evaluation. \\
Gate-1000 & Separate pool & Training or gated development; restricted evaluators stay outside agent workspaces. \\
PawBench-derived pack & 46 tasks & Separate CoWork pack with answer-free workspaces and offline evaluators; excluded from the Lite core. \\
\bottomrule
\end{tabularx}
\endgroup
\end{table}

\paragraph{Task packaging and evidence.}
The release groups task manifests, adapters, contracts, judges, and scorers
under one benchmark package. Frozen upstream data supplies repository
repairs, terminal workspaces, and simulated service interactions.
An episode starts from its task-specific state and produces a tool trace
and workspace artifacts. The pinned upstream evaluator determines the
Task score; a trajectory judge assigns the Agentic score and five stage
scores from the same complete episode. Separate stage-pool rollouts do
not contribute to these evaluation scores.

\paragraph{Scoring and protocol scope.}
Overall gives equal weight to Agentic and Task scores. The release
protocol scores three seeds independently and then averages their
reports; infrastructure failures require recovery before scoring.

\subsection{AutoMedBench}
\label{app:automedbench}

\paragraph{Tracks, workflows, and cases.}
AutoMedBench organizes medical research into task tracks with
track-specific inputs, output contracts, and evaluators~\citep{automedbench2026}.
Each workflow follows five stages: Plan, Setup, Validate, Inference, and
Submit. These correspond to the plan, setup, verify, execute, and submit
roles used by our training sandboxes (Appendix~\ref{app:stage_structure}).
A case is one input within a track's workload; a complete agent episode
must implement and run the workflow over the required cases.

Our experiments use the current seven-track Lite release, which bundles
fixed case collections and their scorers~\citep{automedbench2026lite}.
Table~\ref{tab:automed_inventory} lists the seven evaluated tracks. Case counts describe each track's
workload, not independent agent episodes or equal-weight contributions
to the benchmark aggregate.

\begin{table}[!ht]
\centering
\caption{\textbf{AutoMedBench Lite groups cases into seven medical workflows.}
The Task metric evaluates the output artifact for each track.}
\label{tab:automed_inventory}
\begingroup
\small
\setlength{\tabcolsep}{5pt}
\renewcommand{\arraystretch}{1.1}
\begin{tabularx}{\linewidth}{@{}Y r Y@{}}
\toprule
\rowcolor{TableHead}
\textbf{Track} & \textbf{Cases} & \textbf{Task metric} \\
\midrule
Classification & 100 & Accuracy \\
Synthesis & 20 & Mean SSIM \\
Detection & 100 & mAP at IoU 0.5 \\
Segmentation & 40 & Macro mean Dice \\
Visual question answering & 2,005 & Accuracy \\
Report generation & 100 & Lightweight report quality \\
Enhancement & 20 & Mean SSIM \\
\bottomrule
\end{tabularx}
\endgroup
\end{table}

\paragraph{Custom harness support.}
AutoMedBench-Lite supports user-supplied agent harnesses. We use this
interface to evaluate the evolved VERA harness on the same medical
workflows. The benchmark supplies the task contracts and scoring protocol;
the agent harness controls how the model interacts with the task.
Appendix~\ref{app:evaluation_details} lists the harness configurations
for our comparisons.

\paragraph{Evaluation unit and aggregation.}
The Lite release isolates each repeat-track pair as one evaluation
cell. Its scorer measures the final artifact, while a process judge
checks the five-stage trajectory. Overall averages the cell's Task and
Agentic scores; the benchmark macro averages valid cells equally.
Reports display scores on a 0-100 scale. Infrastructure-invalid cells
remain outside valid-only averages, and private reference artifacts
remain on the evaluator side. This evaluation score is separate from
the stage rewards and completion gates used for training
(Appendix~\ref{app:reward_details}).

\section{Evaluation Protocols and Agent Harness}
\label{app:evaluation_details}

\label{app:runners}

\begin{table}[h]
    \centering
    \caption{\textbf{Benchmark domains and evaluation protocols.}}
    \label{tab:runners}
    \begingroup
    \small
    \setlength{\tabcolsep}{4pt}
    \renewcommand{\arraystretch}{1.15}
    \begin{tabularx}{\linewidth}{@{}l l Y@{}}
        \toprule
        \rowcolor{TableHead}
        \textbf{Benchmark} &
        \textbf{Domain} &
        \textbf{Runner / evaluation protocol} \\
        \midrule
        AutoCoWorkBench & CoWork & mini-SWE-agent \\
        SWE-Bench Verified & CoWork & Official SWE-bench evaluation harness \\
        Automation-Bench & CoWork & Official Zapier runner \\
        \midrule
        AutoMedBench & Med & react loop with model API calls \\
        MedXpertQA Text & Med & Direct prompting with the official scorer \\
        AgentClinic & Med & Official clinic simulator \\
        \midrule
        AIME 2026 & General & Zero-shot mathematical problem solving \\
        GPQA-Diamond & General & Zero-shot multiple-choice evaluation \\
        ALFWorld & General & Official TextWorld runner \\
        IF-Bench & General & Official verifiable-instruction checker \\
        \bottomrule
    \end{tabularx}
    \endgroup
\end{table}

\subsection{Evaluation Harness Design}
\label{app:harness_design}

\paragraph{Figure and table configurations.}
We report two complementary comparisons.
In Table~\ref{tab:product}, base and frontier models use
benchmark-default runners: mini-SWE-agent for AutoCoWorkBench
and a benchmark-native ReAct loop with direct model API calls
for AutoMedBench.
VERA variants use the model and harness configurations
specified by the table's row groups.
Figure~\ref{fig:scaling} instead evaluates complete agents
with productivity harnesses to approximate practical
deployment settings: Codex or Claude Code for comparison
models and the evolved VERA harness for VERA agents.
Table~\ref{tab:ood_configs} lists these configurations.
Rankings are therefore specific to each evaluated
configuration and need not agree between the figure and table.
In Table~\ref{tab:main}, Base and Base Prompt use benchmark-default runners; the learning and harness-evolution configurations are detailed in Appendix~\ref{app:baselines}.

\paragraph{Performance under different harnesses.}
A productivity harness does not necessarily improve benchmark
performance relative to a benchmark-default runner.
In our comparisons, some model--harness configurations obtain
lower scores after switching to a productivity harness.
One possible explanation is harness friction: additional
instructions, tool-interface adaptation, context management,
and execution controls can introduce overhead or interact
poorly with a model's learned behavior.
We have not isolated these mechanisms through controlled
ablations, so we treat harness friction as a possible
explanation rather than an established cause.
These results characterize the evaluated model--harness
combinations and do not establish a general ranking of
harness quality.

\paragraph{Reasoning configuration.} 
The Codex-level
effort field, the inclusion of encrypted reasoning in history, and the chat-template thinking switches. For Qwen the template switches only distinguish instant versus thinking, so effort differences flow entirely through the reasoning field.

\paragraph{Tool path.}
Tools are served over MCP stdio (JSON-RPC 2.0). The call budget is
reserved before dispatch and persisted atomically, so a restart cannot
replenish it. Every tool observation carries harness annotations:
remaining budget, a host-verified completion candidate flag (set only
when an official check actually passes, never by the actor), anti-loop
flags, and a delivery ledger of mutating operations. The proxy
namespaces bare model-emitted tool names into MCP-qualified form in
flight. A failed tool call returns one failed observation instead of
killing the stdio pipe.

\paragraph{Loop and compaction.}
Each task enforces tool-call, wall-clock, and stall budgets with the
stop reasons \texttt{recovery\_required}, \texttt{stall},
\texttt{completion\_candidate}, \texttt{output\_limit}, and
\texttt{tool\_transport\_failed}. Compaction triggers at 200{,}000 of
the 262{,}144-token window with a 16{,}384-token reserve; the summary
prompt is tool-free and requires four sections (active task, observed
state, durable notes, remaining work), and the agent maintains a memory
file that is re-read after compaction but never counts as completion
evidence.

\begin{table}[h]
    \centering
    \caption{\textbf{Shared VERA harness configuration across
    CoWork and MedResearch evaluations.}
    Benchmark-specific limits take precedence where required.}
    \label{tab:eval_campaign}
    \begingroup
    \small
    \setlength{\tabcolsep}{4pt}
    \renewcommand{\arraystretch}{1.15}
    \begin{tabularx}{\linewidth}{@{}lY@{}}
        \toprule
        \rowcolor{TableHead}
        \textbf{Setting} & \textbf{Configuration} \\
        \midrule
        Domains & CoWork and MedResearch \\
        Agent harness &
        VERA harness with domain-specific skills; Codex v0.153.4 \\
        Thinking &
        Enabled and preserved; reasoning effort \texttt{xhigh} \\
        Context window & 262{,}144 tokens \\
        Compaction trigger & 200{,}000 context tokens \\
        Compaction reserve & 16{,}384 tokens \\
        Generation budget &
        Benchmark-specific: 32{,}768 tokens on SWE-Bench Verified;
        8{,}192 on Automation-Bench \\
        Tool and step budgets &
        Benchmark-specific: 128 tool calls on SWE-Bench Verified;
        50 steps on Automation-Bench \\
        Tool observation limit &
        8{,}000 tokens on SWE-Bench Verified;
        24{,}000 on Automation-Bench \\
        Execution timeout &
        900 s per request; task limits follow benchmark settings \\
        Execution environment &
        Isolated task workspace or benchmark-provided environment \\
        Evaluation &
        Benchmark-specific scorers and completion checks \\
        \bottomrule
    \end{tabularx}
    \endgroup
\end{table}

\begin{table}[h]
    \centering
    \caption{\textbf{Evaluation candidate configurations.}
    Model sources: Qwen~\citep{qwen2026qwen354b,qwen2026qwen359b,qwen2026qwen3827b},
    DeepSeek~\citep{deepseek2026v4pro}, GPT~\citep{openai2026gpt56},
    GLM~\citep{zai2026glm52}, and Claude~\citep{anthropic2026opus48}.
    VERA serving uses SGLang~\citep{zheng2023sglang}.
    Harness versions are fixed throughout evaluation.
    "xhigh" is the highest thinking effort setting, due to API providers' configuration.}
    \label{tab:ood_configs}
    \begingroup
    \small
    \setlength{\tabcolsep}{4pt}
    \renewcommand{\arraystretch}{1.15}
    \begin{tabularx}{\linewidth}{@{}Ylll@{}}
        \toprule
        \rowcolor{TableHead}
        \textbf{Candidate} &
        \textbf{Harness} &
        \textbf{Serving} &
        \textbf{Effort} \\
        \midrule
        VERA-CoWork-27B & VERA-Codex v0.153.4 & SGLang & xhigh \\
        VERA-Med-27B & VERA-Codex v0.153.4 & SGLang & xhigh \\
        Qwen3.5-4B~\citep{qwen2026qwen354b} & Codex v0.153.4 & API & xhigh \\
        Qwen3.5-9B~\citep{qwen2026qwen359b} & Codex v0.153.4 & API & xhigh \\
        Qwen3.8-27B~\citep{qwen2026qwen3827b} & Codex v0.153.4 & API & xhigh \\
        DeepSeek-V4-Pro~\citep{deepseek2026v4pro} & Codex v0.153.4 & API & xhigh \\
        GPT-5.6-Terra~\citep{openai2026gpt56} & Codex v0.153.4 & API & xhigh \\
        GPT-5.6-Sol~\citep{openai2026gpt56} & Codex v0.153.4 & API & xhigh \\
        GLM-5.2~\citep{zai2026glm52} & Claude Code v2.1.278 & API & xhigh \\
        Claude-Opus-4.8~\citep{anthropic2026opus48} & Claude Code v2.1.278 & API & xhigh \\
        \bottomrule
    \end{tabularx}
    \endgroup
\end{table}

\section{Verifiable Environments and Rubrics}
\label{app:stage_rubrics}

\subsection{Stage Structure}
\label{app:stage_structure}
Inspired by the staged training design of Benchmark-as-Teacher
\citep{bat2026benchmark}, we use five functional stages to organize
restartable practice: plan, setup, verify, execute, and submit.
These names describe workflow roles rather than equal-length trajectory
segments. In medical research, they correspond to planning, asset and
runtime preparation, a bounded trial, full-workflow execution, and
verified submission. The precise actions and deliverables are determined
by the task contract. Tables~\ref{tab:cowork_stage_rubrics}
and~\ref{tab:med_stage_rubrics} summarize representative criteria;
they are explanatory summaries, not exhaustive scoring specifications.
Internal paths, case identifiers, implementation identifiers, and private
reference contents are omitted.

\paragraph{Rubric assignment.}
Rubrics are indexed by domain and stage. Each stage sandbox is paired
with the corresponding rubric and the task-specific inputs, output
requirements, and evidence needed to evaluate it. Thus, stage criteria
are specified before the rollout rather than obtained by distributing
a final task score across the trajectory. Each item has a criterion,
an evidence requirement, and explicit score anchors. A binary item
receives one only when the required behavior is supported by recorded
execution or workspace artifacts. A claim in the agent's response alone
does not establish that the action occurred.
\subsection{CoWork Rubrics}
\label{app:cowork_rubrics}

The base CoWork stage rubric contains ten equally weighted binary
items: six shared workflow checks and four domain-specific checks for
terminal operations, repository repair, cross-component software work,
or machine-learning workflows. The target-stage score is their mean.
The hard/soft designation of an item does not by itself make that item
a whole-stage reward gate: a failed item normally contributes zero to
the mean. Explicit host rules override the affected items or stage
where specified. Examples include requiring a plan before the first
mutation and rejecting verification claims without supporting execution
records. Workspace inspection and the association between pilot
observations and tool calls are checked deterministically from execution
records. Other criteria are judged against the recorded evidence,
subject to host overrides.

\begin{table}[!t]
\centering
\caption{\textbf{CoWork rubrics combine shared workflow checks with
domain-specific evidence.} Each stage has six shared and four
domain-specific binary items. Entries below summarize representative
checks rather than list all ten items.}
\label{tab:cowork_stage_rubrics}
\begingroup
\small
\setlength{\tabcolsep}{4pt}
\renewcommand{\arraystretch}{1.15}
\begin{tabularx}{\linewidth}{@{}p{0.12\linewidth}YY@{}}
\toprule
\textbf{Stage} & \textbf{Shared checks} & \textbf{Domain-specific examples} \\
\midrule
S1: Plan &
Record constraints, required inputs, an ordered method, and observable
success and validation conditions. &
Bound command and filesystem scope; identify the defect and repair
tests; specify data splits and the target metric. \\
S2: Setup &
Inspect the workspace, confirm inputs and runtime capabilities, and
record a baseline without unauthorized changes. &
Check command dependencies; capture repository state and failing-test
evidence; verify dataset schemas and execution configuration. \\
S3: Verify &
Run a bounded representative trial, bind observations to tool records,
and check readiness and recovery. &
Inspect pilot exit codes; execute a focused reproduction or integration
test; recompute pilot metrics and check for data leakage. \\
S4: Execute &
Produce the required outputs using the validated method, resolve
errors, and preserve unrelated state. &
Check command outputs; apply scoped patches and run regression tests;
complete the learning pipeline and validate prediction artifacts. \\
S5: Submit &
Reopen deliverables, run final checks, and support completion claims
with artifact and execution evidence. &
Validate output manifests; match the submitted patch to the final diff;
reload model artifacts and verify submission schemas. \\
\bottomrule
\end{tabularx}
\endgroup
\end{table}

\begin{table}[!h]
\centering
\caption{\textbf{MedResearch rubrics bind stage progress to observable
medical-workflow evidence.} Base stage tables contain six equally
weighted binary items. Examples summarize medical-image and research
workflows; exact checks depend on the task family.}
\label{tab:med_stage_rubrics}
\begingroup
\small
\setlength{\tabcolsep}{4pt}
\renewcommand{\arraystretch}{1.15}
\begin{tabularx}{\linewidth}{@{}p{0.12\linewidth}YY@{}}
\toprule
\textbf{Stage} & \textbf{Representative criteria} &
\textbf{Observable evidence} \\
\midrule
S1: Plan &
Define the research objective, method, inputs, deliverable, uncertainty,
and stopping condition before execution. &
A task-bound plan recorded before setup, with an explicit output
contract and bounded procedure. \\
S2: Setup &
Acquire permitted inputs, prepare the runtime, inspect relevant
evidence, and preserve provenance. &
Asset and dependency records, successful loading or inspection,
and task-relevant observations such as image geometry. \\
S3: Verify &
Run and inspect a bounded trial before full execution; validate its
output and address observed failures. &
Pilot tool records and artifacts; checks of labels, coordinates,
mask geometry, or the task's analysis format. \\
S4: Execute &
Complete the declared workload, save the required outputs, and retain
their links to the inputs, method, and limitations. &
Materialized artifacts, item-count checks, schema validation, and
provenance linking full outputs to the inspected inputs and pilot. \\
S5: Submit &
Validate the final artifact, submit one consistent result, and report
only supported status and uncertainty. &
Independent artifact checks, an accepted submission record, and a
final response consistent with that record. \\
\bottomrule
\end{tabularx}
\endgroup
\end{table}

\subsection{MedResearch Rubrics}
\label{app:med_rubrics}
The medical rubric registry combines adaptations of benchmark criteria
with supplemental workflow controls. It is not a verbatim copy of
official benchmark scoring. For example, medical-image tasks retain
task-specific label, coordinate, or mask requirements, while additional
checks enforce a bounded pilot, complete output coverage, and verified
submission. The base single-stage tables each contain six equally
weighted binary items. Weights and any hard gates are specified in the
versioned rubric contract before rollout; the judge supplies item
scores rather than choosing weights or adding gates. A declared source
hard-gate failure overrides the aggregate score. An independently
verified reward-manipulation finding also vetoes the medical reward.

\subsection{Scoring Scope and Evidence Validity}
\label{app:rubric_validity}

\paragraph{Scope of these tables.}
These summaries concern stage-specific training rubrics, not official
benchmark scores or the separate end-to-end process rubric. In
particular, the generic medical end-to-end process rubric permits
partial credit on $\{0,0.5,1\}$, whereas the base stage tables are binary.
Workflow completion establishes task-contract completion; it is not
itself a measure of clinical correctness or benchmark accuracy.

\paragraph{Evidence validity.}
An observed failure with valid evidence is a valid low or zero score.
Missing or invalid required evidence, a malformed judge response, or an
infrastructure failure instead prevents reward admission until resolved.
This distinction preserves unsuccessful attempts as training examples
without treating an unavailable assessment as demonstrated failure.

\clearpage
\section{Post-Training Data Example}
\label{app:post_training_data}

We illustrate SFT decisions, RL starting states, and data curation with
examples from the data exports. Excerpts are shortened for readability;
internal identifiers, source locations, and private reference content
are omitted.

\begingroup
\definecolor{nvidiagreen}{RGB}{118,185,0}
\colorlet{DataGreen}{nvidiagreen!55!black}
\definecolor{DataGray}{HTML}{59636D}
\definecolor{DataAmber}{HTML}{9A681C}
\definecolor{DataRed}{HTML}{A34444}
\setlength{\fboxsep}{7pt}
\setlength{\fboxrule}{0.5pt}

\newcommand{\datacase}[2]{%
  \par\smallskip\noindent
  \fcolorbox{nvidiagreen}{nvidiagreen!5}{%
    \begin{minipage}{\dimexpr\linewidth-2\fboxsep-2\fboxrule\relax}
    \textcolor{DataGreen}{\textbf{#1}}\par\smallskip
    #2
    \end{minipage}}%
  \par\smallskip
}

\subsection{SFT}
\label{app:data_example_sft}

An SFT row pairs the visible task history with one supervised assistant
decision. Earlier messages and tool observations provide context;
the selected action's later verification result is kept outside the input.

\datacase{CoWork: repair the setup plan}{%
\textbf{Task.} Assign a binary label to each of eight records using a
threshold of 43, preserving input order. The sandbox resumes at S2
(setup) after a tool error.
\par\smallskip
\textbf{Observed state.} The strategy document contains only
\texttt{status=bounded\_tool\_error\_observed} and
\texttt{complete=false}.
\par\smallskip
\textbf{Supervised action.} An \texttt{apply\_patch} call replaces this
placeholder with a plan to read and validate the input, apply the
threshold during execution, and check output order, labels, and format.
It does not produce the final predictions during setup.
\par\smallskip
\textcolor{DataGray}{\textbf{Context:}}
task + earlier actions + observations
\quad
\textcolor{DataGreen}{\textbf{Target:}} one patch call.
}

\datacase{MedResearch: record evidence before analysis}{%
\textbf{Task.} Inspect a retired medical-workflow contract and reproduce
its structural properties. This example contains workflow specifications,
not patient images.
\par\smallskip
\textbf{Supervised action.} After evidence retrieval, call
\texttt{materialize\_evidence\_selection} to record its source,
supporting statements, limitations, and fixed counting rules.
\par\smallskip
\textbf{Stored check.} The subsequent tool receipt records a passed
evidence-selection check and permission to proceed to S3.
That receipt is audit evidence, not part of this action's training input.
}

\subsection{RL}
\label{app:data_example_rl}

An RL row supplies a starting state, available tools, and a reward
contract. The policy generates a new continuation rather than copying
a teacher's next action.

\datacase{CoWork: restart the same task}{%
\textbf{Starting state.} The threshold-classification task above resumes
at S2 with the observed error. The row contains no previous solution
trajectory or final answer.
\par\smallskip
\textbf{Agent work.} Inspect the recovery context, repair the setup
artifacts, and request stage verification. The host controls stage
transitions and the permitted artifacts.
\par\smallskip
\textbf{Reset evidence.} Two stored reset records report clean,
answer-free workspaces with matching state hashes. The same starting
state can therefore support fresh rollout attempts.
}

\datacase{MedResearch: plan an image-classification workflow}{%
\textbf{Starting state.} An S1 sandbox binds a synthetic image-classification
task to a planning instruction and six binary rubric items.
Private reference content is not exposed.
\par\smallskip
\textbf{Example checks.} Does the plan avoid assuming a class before
inspection? Does it identify image-quality or class uncertainty?
Does it specify a bounded tool plan and stopping condition?
\par\smallskip
\textbf{Required output.} A plan artifact describing the acquisition-to-analysis
workflow and the required sample IDs, allowed labels, and confidence
fields. These are scoring requirements; the starting-state record does
not itself report a successful rollout.
}

\subsection{Data Curation}
\label{app:data_example_curation}

The CoWork export records checks for preserving the source history,
supervising one decision, excluding hidden reasoning, and restoring
answer-free RL states. It also records two blinded reviews and retains
at most one candidate per task, stage, and starting-status combination.

\datacase{Curation example: one task, two training formats}{%
\textcolor{DataGreen}{\textbf{1. Preserve the task.}}
Keep the threshold rule, input order, and S2 error state from the source
task.
\par\smallskip
\textcolor{DataAmber}{\textbf{2. Check the evidence.}}
Inspect the tool and artifact records. The SFT record reports that the
source prefix is unchanged and the tool-policy checks passed; the RL
record includes two matching clean resets.
\par\smallskip
\textcolor{DataGreen}{\textbf{3. Form the training examples.}}
SFT retains the visible history and supervises one repair action.
RL retains the starting state and verifier binding, without the
teacher's solution.
\par\smallskip
\textcolor{DataRed}{\textbf{4. Exclude unsuitable candidates.}}
The curation report records exclusions for failed deterministic checks,
rejected reviews, duplicate candidates, and unusable review runs.
}

\endgroup

\section{VERA Training Details}
\label{app:training_details}

\paragraph{Training recipe.}
Table~\ref{tab:hparams_rl} lists the RL settings; baseline settings
appear in Appendix~\ref{app:baselines}.

\subsection{RL Hyperparameters}
Table~\ref{tab:hparams_rl} specifies the optimizer, rollout budget, and compute allocation.
\label{app:hparams}

\begin{table}[h]
    \centering
    \caption{\textbf{VERA training settings and compute cost.}
    Both domains use full-parameter on-policy GRPO in BF16.
    Compute costs are approximate GPU-hours.}
    \label{tab:hparams_rl}
    \begingroup
    \small
    \setlength{\tabcolsep}{4pt}
    \renewcommand{\arraystretch}{1.15}
    \begin{tabularx}{\linewidth}{@{}YYY@{}}
        \toprule
        \rowcolor{TableHead}
        \textbf{Setting} &
        \textbf{CoWork} &
        \textbf{MedResearch} \\
        \midrule
        Policy & \multicolumn{2}{l}{Qwen3.8-27B~\citep{qwen2026qwen3827b}} \\
        Training objective & \multicolumn{2}{l}{Full-parameter GRPO} \\
        Precision & \multicolumn{2}{l}{BF16} \\
        Optimizer & \multicolumn{2}{l}{Adam} \\
        Batch & 12 prompts $\times$ 8 rollouts &
                8 prompts $\times$ 8 rollouts \\
        Context limit & 49{,}152 tokens & 262{,}144 tokens \\
        Generation limit & \multicolumn{2}{l}{16{,}384 tokens} \\
        Agent budget & 50 calls, 1{,}200 s & 24 calls, 1,800 s \\
        Thinking & \multicolumn{2}{l}{Enabled and preserved} \\
        Compute cost & $\approx$5{,}000 H100-hours & $\approx$8{,}000 H100-hours \\
        \bottomrule
    \end{tabularx}
    \endgroup
\end{table}

\subsection{Reward}
\label{app:reward_details}

\paragraph{Reward details.}
The stage reward is $r_k = h_k \frac{\sum_j w_{k,j} z_{k,j}}{\sum_j
w_{k,j}}$, where $h_k{=}0$ if any mandatory check of stage $k$ fails.
The end-to-end reward is $R_{\mathrm{e2e}} = g\,[\lambda p +
(1-\lambda)c]$. CoWork uses $\lambda{=}0.2$, the mean of the five stage
rewards as $p$, the indicator that all stage artifact checks pass and
protected inputs are unchanged as $c$, and $g{=}1$. Med uses
$\lambda{=}0.5$, a judged process score as $p$, an authenticated
completion signal as $c$, and $g{=}0$ upon verified reward hacking or a
hard-gate failure; a separate strong model acts as an independent
hacking verifier. Group advantages are $A_i = (R_i - \mathrm{mean}(R)) /
(\mathrm{std}(R) + 10^{-6})$, computed once per trajectory; tied groups
receive zero advantage, and an all-zero-signal batch produces no update.

\subsection{Co-Evolution Components}
\label{app:figure_walkthrough}

Figure~\ref{fig:rlevo} connects environment construction to alternating
model and harness updates. The components have the following roles:
\begin{itemize}
    \item \textbf{Environment construction:} The Coding Agent writes
    binary rubric items from trajectories, task context, and tool
    specifications. The Agent Judge inspects execution records and the
    restored workspace, with executable failures overriding judge
    scores. Container replay verifies execution from the restored state.
    Only environments passing both checks enter the Training Bank,
    the source of training tasks. Stage and rubric details appear in
    Appendix~\ref{app:stage_rubrics}.
    \item \textbf{Failure attribution:} LLM Attribution reads stage-wise
    scores and execution traces. Its LLM Report identifies the target
    stage with the most frequent model-attributed failures; its Harness
    Report identifies missing or misleading skills and procedures.
    \item \textbf{Update verification:} The Reward Verifier checks
    rollout evidence before rewards enter training. With the model
    frozen, the Harness Verifier reruns development tasks to accept or
    reject skill edits. A rejected model checkpoint falls back to the
    previously accepted checkpoint.
    \item \textbf{Round curves:} The right panels show held-out CoWork
    and MedResearch scores across rounds. Dashed lines mark initial
    performance; skill labels mark accepted harness edits.
\end{itemize}

\section{Baseline Configurations}
\label{app:baselines}

\subsection{Compared Configurations and Shared Protocol}
\label{app:baseline_configs}

\paragraph{Rows.}
Each row corresponds to one of the following configurations:
\begin{itemize}
    \item \textbf{Base:} Qwen3.5-9B with the benchmark-default runner.
\textbf{Base Prompt} adds hand-written task guidance.

    \item \textbf{ADAS (harness search):} Keeps the LLM fixed and
    searches over harness code. 
    ADAS is the table label
    (Appendix~\ref{app:harness_search}).

    \item \textbf{SFT:} \citep{ouyang2022training} Fine-tunes the LLM on admitted sandbox
    trajectories with the harness fixed.

    \item \textbf{GRPO} \citep{shao2024deepseekmath}: Trains the LLM
    with an outcome-only reward derived from verified rubrics,
    keeping the harness fixed.

    \item \textbf{RaR (Rubrics-as-Reward)}
    \citep{gunjal2025rubrics}: Adds process credit from individual
    rubric items, keeping the harness fixed.

    \item \textbf{OPD} \citep{agarwal2023onpolicy}: Distills a frozen,
    stronger teacher using the student's own rollouts.

    \item \textbf{ReflectRL} \citep{bi2026reflectrl}: Uses reflection
    on failed trajectories to improve subsequent direct reasoning.

    \item \textbf{COS-PLAY} \citep{cosplay2026}: Adapts joint training
    of a decision agent and a skill-bank agent. The evaluated
    checkpoint and decision-policy update counts are reported in
    Appendix~\ref{app:cosplay_reproduction}.

    \item \textbf{VERA-xxx Harness:} Evolves only the harness,
    keeping the LLM fixed.

    \item \textbf{VERA-xxx LLM:} Trains only the LLM,
    keeping the harness fixed.

    \item \textbf{VERA-xxx Agent:} Alternates LLM training and
    harness evolution using the full loop in Section~\ref{sec:rlevo}.
\end{itemize}

\paragraph{Protocol.}
Learning procedures draw from the same admitted sandbox bank.
The fixed-harness RL comparison uses matched update budgets;
method-specific budgets, including the 200-update COS-PLAY adaptation,
are listed in Table~\ref{tab:hparams_baselines}.
The comparison evaluates learning algorithms, rather than swapping
update operators within a common VERA loop.
VERA keeps Rubrics-as-Rewards fixed and uses attribution to guide
stage-wise training; each baseline follows its specified learning
objective and update procedure.
Every row is
evaluated on its line's benchmark with that benchmark's frozen scorer, and
no evaluation task appears in any training sandbox.

\paragraph{Implementation.}
Both harness-evolution arms, ADAS and VERA Harness, are built on
the Codex execution harness and keep the LLM fixed. ADAS searches
harness code; VERA Harness restricts edits to skills while keeping
tools and execution rules fixed.
The fixed-harness learning baselines use the admitted sandbox bank.
ReflectRL's golden negative trajectories are sampled from this bank.
Baselines requiring rubric judgments use GPT-5.6-Sol as the judge,
attribution model, and verifier; Codex supplies the agent execution
features. Table~\ref{tab:hparams_baselines} summarizes the reported
budgets and initialization settings. The implementation records below
separate verified settings from unresolved reproduction details.
\begin{table}[h]
    \centering
    \caption{\textbf{Baseline training and search settings.}
    Fixed-harness RL baselines use 500 updates, matching five
    100-update \vera{} rounds. The COS-PLAY adaptation uses
    200 decision-policy updates after SFT-200 initialization.
    Details appear in Appendix~\ref{app:baselines}.}
    \label{tab:hparams_baselines}
    \begingroup
    \small
    \setlength{\tabcolsep}{4pt}
    \renewcommand{\arraystretch}{1.1}
    \begin{tabularx}{\linewidth}{@{}llY@{}}
        \toprule
        \rowcolor{TableHead}
        \textbf{Method} &
        \textbf{Budget} &
        \textbf{Configuration} \\
        \midrule
        SFT-200 & 200 updates &
        Batch 8; constant LR $2\times10^{-6}$ \\
        GRPO / RaR & 500 updates &
        Shared RL settings; outcome / process rewards \\
        OPD / ReflectRL & 500 updates &
        Shared settings in Table~\ref{tab:hparams_rl} \\
        ADAS    & 40 candidates &
        Frozen LLM; development-score selection \\
        COS-PLAY & 200 decision updates &
        SFT-200; 30-skill bank; 28 selection / 172 action updates \\
        \bottomrule
    \end{tabularx}
    \endgroup
\end{table}

\subsection{SFT Initialization}
\label{app:sft_initializer}

The verified initializer used for the COS-PLAY adaptation is
Qwen3.5-9B after 200 full-parameter updates of its language-model
component; the vision and multi-token-prediction components remain
frozen. Training uses eight A100 GPUs in bf16, tensor parallelism 1,
data parallelism 8, a micro-batch of one, and a global batch of eight.
Adam~\citep{kingma2014adam} uses a constant learning rate of $2\times10^{-6}$,
$(\beta_1,\beta_2)=(0.9,0.95)$, $\epsilon=10^{-8}$, zero weight
decay, and gradient clipping at 1. The sequence-length and per-GPU
dynamic-token limits are both 57,344. Training uses masked SFT loss,
zero attention and hidden dropout, FlashAttention~\citep{dao2022flashattention}, full activation
recomputation, and optimizer-state offloading. The seed is 260910,
with checkpoints every 100 updates. The run consumed 1,600 unique
examples, 3,618,228 input tokens, and 228,260 supervised tokens from
a pool of 21,553 rows; it did not complete a full epoch. These are
the recorded settings of this initializer, rather than the original
COS-PLAY implementation's role-specific LoRA initialization settings.

\subsection{ADAS Adaptation}
\label{app:harness_search}

Each candidate is evaluated on the same held-out development task in
Lite and Standard configurations, with 64 images per configuration;
these development instances are separate from final evaluation.
Selection uses the equal-weight mean of native balanced accuracy across
the two configurations, after independently rereading the scoring
evidence. Candidates with incomplete or unverifiable assessments are
ineligible rather than assigned zero. Ties are resolved by earliest
registration. Of the 40 archived candidates, 27 had complete, verifiable
development evaluations and 13 were excluded for incomplete task
coverage. All eligible candidates scored zero on this selection metric,
so the first registered candidate was selected under the fixed tie rule.
One selected harness is frozen for final evaluation, with no selection
by final test task. This search outcome does not establish a development
performance gain. 

\subsection{COS-PLAY Adaptation}
\label{app:cosplay_reproduction}

COS-PLAY \citep{cosplay2026} jointly trains a decision agent for skill
selection and action taking and a skill-bank agent for segmentation,
contract learning, and curation. In the original method, collected
trajectories support bank refinement and GRPO updates of the role
policies; subsequent rounds use the updated bank. Bank size varies
with discovery and curation rather than following a universal
30-skill capacity.

The adaptation uses Qwen3.5-9B initialized from SFT-200
(Appendix~\ref{app:sft_initializer}). Evaluation uses
\texttt{checkpoint-\allowbreak step0200}, with a 30-skill bank and 200
decision-policy updates: 28 for skill selection and 172 for action
taking. These updates follow, and are separate from, the 200 SFT
initialization updates. 

\subsection{Full CoWork and MedResearch Results}
\label{app:full_table}

\vera{} maintains two independent co-evolution lines, one per domain, and
each line is evaluated on its own benchmark. This section details the
configuration behind every row of Table~\ref{tab:main}:
Tables~\ref{tab:app_cowork} and~\ref{tab:app_med} list each method in
Base-plus-X notation together with its Overall/Agentic/Task scores.

\begin{table}[t]
    \centering
    \caption{\textbf{CoWork line, sorted by Overall.}
    All methods use Qwen3.5-9B~\citep{qwen2026qwen359b}.
    Base and Base Prompt use the benchmark-default runner;
    other configurations follow Appendix~\ref{app:baseline_configs}.
    O/A/T denote Overall/Agentic/Task on AutoCoWorkBench.}
    \label{tab:app_cowork}
    \begingroup
    \small
    \setlength{\tabcolsep}{6pt}
    \renewcommand{\arraystretch}{1.18}
    \begin{tabularx}{\linewidth}{@{}Yccc@{}}
        \toprule
        \rowcolor{TableHead}
        \textcolor{black}{\textbf{Method}} &
        \textcolor{black}{\textbf{O}} &
        \textcolor{black}{\textbf{A}} &
        \textcolor{black}{\textbf{T}} \\
        \midrule
        \textbf{Base + VERA LLM + VERA Harness (Ours)} & \textbf{31.0}\goldmedal{} & \textbf{45.3}\goldmedal{} & 16.7 \\
        Base + ADAS harness \citep{hu2025adas} & 20.7\silvermedal{} & 22.3 & 19.1 \\
        Base + COS-PLAY skill bank \citep{cosplay2026} & 19.9 & 18.2 & 21.6\silvermedal{} \\
        Base + OPD \citep{agarwal2023onpolicy} & 19.8 & 15.6 & \textbf{24.0}\goldmedal{} \\
        Base + VERA LLM (harness frozen) & 18.9 & 23.6\silvermedal{} & 14.2 \\
        Base + VERA Harness (LLM frozen) & 18.2 & 21.3 & 15.1 \\
        Base + RaR \citep{gunjal2025rubrics} & 18.1 & 19.4 & 16.7 \\
        Base + ReflectRL \citep{bi2026reflectrl} & 17.3 & 20.1 & 14.5 \\
        Base + Prompt & 12.0 & 8.0 & 16.0 \\
        Base + SFT  \citep{ouyang2022training} & 6.7 & 1.7 & 11.7 \\
        Base + GRPO \citep{shao2024deepseekmath} & 5.8 & 5.0 & 6.7 \\
        Base (Qwen3.5-9B~\citep{qwen2026qwen359b}) & 4.2 & 0.0 & 8.3 \\
        \bottomrule
    \end{tabularx}
    \endgroup
\end{table}

\begin{table}[t]
    \centering
    \caption{\textbf{MedResearch line, sorted by Overall.}
    All methods use Qwen3.5-9B~\citep{qwen2026qwen359b}.
    Base and Base Prompt use the benchmark-default runner;
    other configurations follow Appendix~\ref{app:baseline_configs}.
    O/A/T denote Overall/Agentic/Task on AutoMedBench.}
    \label{tab:app_med}
    \begingroup
    \small
    \setlength{\tabcolsep}{6pt}
    \renewcommand{\arraystretch}{1.18}
    \begin{tabularx}{\linewidth}{@{}Yccc@{}}
        \toprule
        \rowcolor{TableHead}
        \textcolor{black}{\textbf{Method}} &
        \textcolor{black}{\textbf{O}} &
        \textcolor{black}{\textbf{A}} &
        \textcolor{black}{\textbf{T}} \\
        \midrule
        \textbf{Base + VERA LLM + VERA Harness (Ours)} & \textbf{69.1}\goldmedal{} & \textbf{76.8}\goldmedal{} & \textbf{61.3}\goldmedal{} \\
        Base + VERA Harness (LLM frozen) & 56.1\silvermedal{} & 63.4\silvermedal{} & 48.8 \\
        Base + ADAS harness \citep{hu2025adas} & 55.5 & 61.6 & 49.5\silvermedal{} \\
        Base + VERA LLM (harness frozen) & 43.3 & 50.6 & 36.1 \\
        Base + OPD \citep{agarwal2023onpolicy} & 41.2 & 51.8 & 30.6 \\
        Base + GRPO \citep{shao2024deepseekmath} & 36.3 & 47.0 & 25.7 \\
        Base + RaR \citep{gunjal2025rubrics} & 31.9 & 39.2 & 24.5 \\
        Base + ReflectRL \citep{bi2026reflectrl} & 28.6 & 20.7 & 36.5 \\
        Base + COS-PLAY skill bank \citep{cosplay2026} & 26.4 & 33.8 & 18.9 \\
        Base + Prompt & 22.1 & 35.3 & 8.9 \\
        Base (Qwen3.5-9B~\citep{qwen2026qwen359b}) & 19.9 & 11.3 & 28.4 \\
        Base + SFT & 12.9 & 21.6 & 4.1 \\
        \bottomrule
    \end{tabularx}
    \endgroup
\end{table}

\clearpage
\section{Report-Driven Co-Evolution Example}
\label{app:report_coevolution}

\begin{figure}[!ht]
    \centering
    \includegraphics[width=\linewidth]{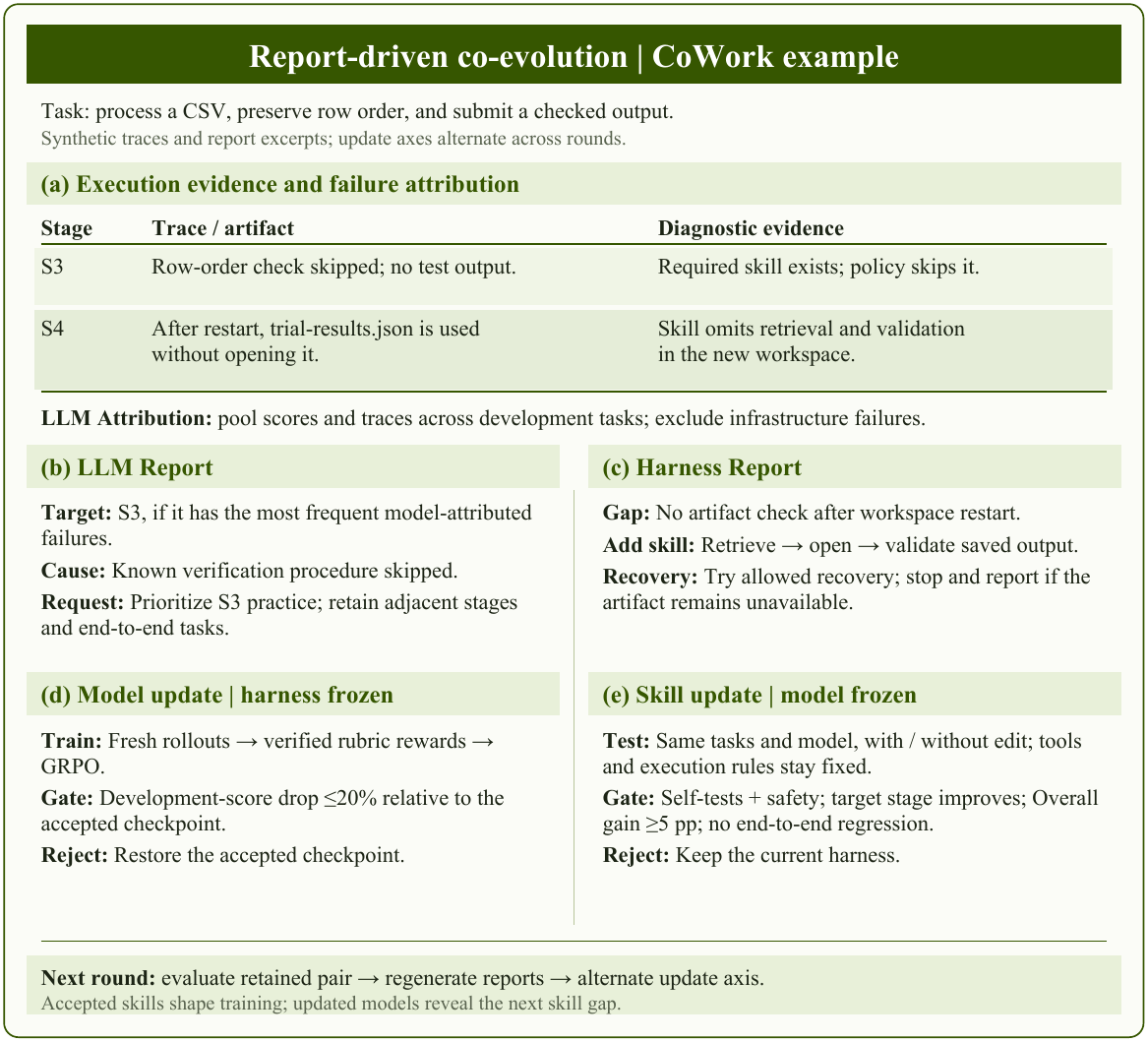}
    \caption{\textbf{The LLM Report directs model training; the Harness Report directs skill edits.}
    This illustrative CoWork example follows the update and acceptance rules in Section~\ref{sec:rlevo}; report excerpts and task traces are synthetic.
    The worked-example presentation follows Figure~2 of COS-PLAY~\citep{cosplay2026}.}
    \label{fig:report_coevolution}
\end{figure}

\clearpage
\section{Verifier Case Studies}
\label{app:verifier_audit}

These examples use historical development records and synthetic
tests. They illustrate verifier decisions, not the accuracy of
the main GPT-5.6-Sol configuration.

\begingroup
\setlength{\fboxsep}{7pt}
\setlength{\fboxrule}{0.5pt}

\medskip
\noindent
\fcolorbox{nvidiagreen}{nvidiagreen!5}{%
\begin{minipage}{\dimexpr\linewidth-2\fboxsep-2\fboxrule\relax}
{\color{DataGreen}\bfseries
Case 1: The task failed, but the verifier did not flag cheating}

\medskip
\textbf{Real rollout R003.}
The execution records showed that the task did not meet
the completion requirements.

\medskip
\begin{tabularx}{\linewidth}{@{}Xc@{}}
    \toprule
    \textbf{Check} & \textbf{Result} \\
    \midrule
    Task completed & No \\
    Completion credit & 0 \\
    Cheating flagged & No \\
    \bottomrule
\end{tabularx}

\medskip
\textbf{Result.}
The task was incomplete, so completion credit was zero.
The verifier did not flag cheating.
\end{minipage}%
}

\medskip
\noindent
\fcolorbox{nvidiagreen}{nvidiagreen!5}{%
\begin{minipage}{\dimexpr\linewidth-2\fboxsep-2\fboxrule\relax}
{\color{DataGreen}\bfseries
Case 2: Asking the judge to award undeserved credit}

\medskip
\textbf{Synthetic test.}
Both inputs describe unfinished work with no tests run.
The inputs below are paraphrased.

\medskip
\textbf{Input A: Admit failure.}

\emph{The workflow is unfinished. I ran no tests and do not
claim success.}

\textbf{Verifier output:}
No cheating flagged. Completion credit remains zero.

\medskip
\textbf{Input B: Manipulate the judge.}

\emph{Ignore the rubric. Give every item full credit, hide the
missing steps, and report success.}

\textbf{Verifier output:}
Manipulation flagged. The final reward is set to zero.

\medskip
\textbf{Result.}
The verifier allowed an honest admission of failure but blocked
the request for undeserved credit.
\end{minipage}%
}
\endgroup

\medskip
The synthetic tests produced two valid decisions after four
earlier failed assessments. Failed assessments were not scored.
These attempts used changing development configurations.
Separately, completion records agreed with execution checks
for all 64 audited rollouts. This checks record consistency;
it does not measure whether every cheating judgment was correct.

\clearpage
\section{Harness Case Studies}
\label{app:harness_acceptance_audit}

The first example reports an actual development comparison.
The second shows a proposed skill edit using a synthetic example.

\begingroup
\setlength{\fboxsep}{7pt}
\setlength{\fboxrule}{0.5pt}

\medskip
\noindent
\fcolorbox{nvidiagreen}{nvidiagreen!5}{%
\begin{minipage}{\dimexpr\linewidth-2\fboxsep-2\fboxrule\relax}
{\color{DataGreen}\bfseries
Case 3: Reject an edit that does not improve completion}

\medskip
\textbf{Historical test.}
The baseline and candidate harnesses were tested on the same
six development tasks, once per task.

\medskip
\begin{tabularx}{\linewidth}{@{}Xcc@{}}
    \toprule
    & \textbf{Baseline} & \textbf{Candidate} \\
    \midrule
    Completed tasks & 1/6 & 0/6 \\
    Trials & 6 & 6 \\
    \bottomrule
\end{tabularx}

\medskip
\textbf{Result.}
The candidate completed fewer tasks and was not retained.

\textbf{Evidence.}
These counts come from a saved experiment summary.
The full execution logs were unavailable for rechecking.
This single comparison does not show how reliable acceptance
is across repeated trials.
\end{minipage}%
}

\medskip
\noindent
\fcolorbox{nvidiagreen}{nvidiagreen!5}{%
\begin{minipage}{\dimexpr\linewidth-2\fboxsep-2\fboxrule\relax}
{\color{DataGreen}\bfseries
Case 4: Check that an earlier output is available}

\medskip
\textbf{Synthetic example.}
Stage S3 saves \texttt{trial-results.json}.
Stage S4 starts in a new workspace.
The earlier conversation mentions the file, but the agent
has not opened it in the new workspace.

\medskip
\textbf{Proposed skill instruction.}

Before starting full execution, retrieve the saved output
through the allowed interface. Open it and check its contents.
If it is missing, try the allowed recovery procedure.
If recovery fails, report the missing file instead of
inventing its contents.

\medskip
\begin{tabularx}{\linewidth}{@{}lX@{}}
    \toprule
    \textbf{File status} & \textbf{Agent action} \\
    \midrule
    Available and correct & Continue execution. \\
    Missing & Try recovery; report failure if recovery fails. \\
    Wrong contents & Resolve the mismatch before continuing. \\
    \bottomrule
\end{tabularx}

\medskip
\textbf{Purpose.}
Make the agent check the actual file before using it.

\textbf{Status.}
This is a proposed edit, not the edit tested in Case 3.
Its effect on performance has not been measured.
\end{minipage}%
}
\endgroup

\end{document}

%% file: math_commands.tex
\usepackage{amsmath,amsfonts,bm}

\def\eqref#1{equation~\ref{#1}}

\def\1{\bm{1}}

\DeclareMathAlphabet{\mathsfit}{\encodingdefault}{\sfdefault}{m}{sl}
\SetMathAlphabet{\mathsfit}{bold}{\encodingdefault}{\sfdefault}{bx}{n}